\documentclass[sigconf]{acmart}

\usepackage{latexsym}
\usepackage{multicol}
\usepackage{amsmath}
\usepackage{multirow}
\usepackage{multicol}
\usepackage{hyperref}
\usepackage{array}
\usepackage{float}
\usepackage{arydshln}
\usepackage{enumitem}
\usepackage{amsmath}

\usepackage{amssymb}
\usepackage{subfig}

\usepackage{sidecap}

\newcommand{\dataset}{\texttt{GOTHate}}
\newcommand{\model}{\texttt{HEN-mBERT}}
\newcommand{\datasetfull}{\textbf{G}eo-p\textbf{O}litical \textbf{T}opical \textbf{H}ate dataset}
\newcommand{\modelfull}{\textbf{H}istory-\textbf{E}xamplar-\textbf{N}etwork \textbf{I}nfused m\textbf{B}ERT}
\AtBeginDocument{%
  \providecommand\BibTeX{{%
    \normalfont B\kern-0.5em{\scshape i\kern-0.25em b}\kern-0.8em\TeX}}}

\setcopyright{acmcopyright}
\copyrightyear{2023}
\acmYear{2023}
\setcopyright{acmlicensed}\acmConference[KDD '23]{Proceedings of the 29th ACM SIGKDD Conference on Knowledge Discovery and Data Mining}{August 6--10, 2023}{Long Beach, CA, USA}
\acmBooktitle{Proceedings of the 29th ACM SIGKDD Conference on Knowledge Discovery and Data Mining (KDD '23), August 6--10, 2023, Long Beach, CA, USA}
\acmPrice{15.00}
\acmDOI{10.1145/3580305.3599896}
\acmISBN{979-8-4007-0103-0/23/08}

\ccsdesc[500]{Computing methodologies~Language resources}
\ccsdesc[500]{Computing methodologies~Supervised learning by classification}

\keywords{Online Hate Speech, Hindi-Codemixed Dataset, Benchmark Comparision, Endogenous Signal Modeling, Content Moderation Tool.}

\begin{document}

% \title{Catch-22: When Hate Speech Dataset Mimics \\Real-World Discourse}
\title{Revisiting Hate Speech Benchmarks: From Data Curation to System Deployment}


% }
\settopmatter{printacmref=true}
% \setcopyright{none}
% \renewcommand\footnotetextcopyrightpermission[1]{}

% \affiliation{%
%   \country{The Th{\o}rv{\"a}ld Group}
%   % \streetaddress{1 Th{\o}rv{\"a}ld Circle}
%   % \city{Hekla}
%   % \country{Iceland}
%   }

% \affiliation{%
%   \country{The Th{\o}rv{\"a}ld Group}
%   % \streetaddress{1 Th{\o}rv{\"a}ld Circle}
%   % \city{Hekla}
%   % \country{Iceland}
%   }

% \affiliation{%
%   \country{The Th{\o}rv{\"a}ld Group}
%   % \streetaddress{1 Th{\o}rv{\"a}ld Circle}
%   % \city{Hekla}
%   % \country{Iceland}
%   }

% \affiliation{%
%   \country{The Th{\o}rv{\"a}ld Group}
%   % \streetaddress{1 Th{\o}rv{\"a}ld Circle}
%   % \city{Hekla}
%   % \country{Iceland}
%   }
\author{Atharva Kulkarni$^{1*}$, Sarah Masud$^{2*}$, Vikram Goyal$^2$, Tanmoy Chakraborty$^3$}
 \affiliation{%
   \country{
   $^1$Carnegie Mellon University; $^2$IIIT-Delhi, India; $^3$IIT Delhi, India}
 }
\email{atharvak@cs.cmu.edu, {sarahm,vikram}@iiitd.ac.in, tanchak@iitd.ac.in}
% \renewcommand{\shortauthors}{Sarah Masud, Manjot Bedi, Mohammad Aflah Khan, Md. Shad Akhtar, Tanmoy Chakraborty}
% \renewcommand{\shortauthors}{Sarah Masud et al.}
\renewcommand{\shortauthors}{Kulkarni et al.}

\begin{abstract}
Social media is awash with hateful content, much of which is often veiled with linguistic and topical diversity. The benchmark datasets used for hate speech detection do not account for such divagation as they are predominantly compiled using hate lexicons. However, capturing hate signals becomes challenging in neutrally-seeded malicious content. Thus, designing models and datasets that mimic the real-world variability of hate warrants further investigation. 

To this end, we present \dataset, a large-scale code-mixed crowdsourced dataset of around $51k$ posts for hate speech detection from Twitter. \dataset\ is neutrally seeded, encompassing different languages and topics. We conduct detailed comparisons of \dataset\ with the existing hate speech datasets, highlighting its novelty. We benchmark it with $10$ recent baselines. Our extensive empirical and benchmarking experiments suggest that \dataset\ is hard to classify in a text-only setup. Thus, we investigate how adding endogenous signals enhances the hate speech detection task. We augment \dataset\ with the user's timeline information and ego network, bringing the overall data source closer to the real-world setup for understanding hateful content. Our proposed solution \model, is a modular, multilingual, mixture-of-experts model that enriches the linguistic subspace with latent endogenous signals from history, topology, and exemplars. \model\ transcends the best baseline by $2.5\%$ and $5\%$ in overall macro-F1 and hate class F1, respectively. Inspired by our experiments, in partnership with Wipro AI, we are developing a semi-automated pipeline to detect hateful content as a part of their mission to tackle online harm.\footnote{\color{red}{\noindent\textbf{Disclaimer:} The paper contains samples of hate speech, which are only included for contextual understanding. We tried our best to censor and not support these views.}}
\end{abstract}

\maketitle
\def\thefootnote{*}\footnotetext{Equal Contributions.}\def\thefootnote{\arabic{footnote}}

\section{Introduction}
Due to their democratized nature, social media platforms serve as breeding grounds for the unwarranted spread of hate speech \cite{fortuna2018survey, masud2021hate, DBLP:journals/corr/abs-2201-00961}, opprobrious offensive opinions \cite{zampieri-etal-2019-semeval}, and provocative propaganda \cite{da-san-martino-etal-2019-findings}. Such proliferation of hate speech disrupts the harmony and cohesiveness of online and offline communities \cite{mathew2019spread, bilewicz2020hate}. Thus, the diachronic study of hate speech's digital footprint has been an active research topic. Despite the aggressive work in this domain, we are yet to confront the myriad issues associated with hate speech benchmarking in a consistent manner. 

{\textbf{Limitations of benchmark hate speech datasets:}} The common vogue for curating hate speech datasets is using hate lexicons \cite{waseem-hovy-2016-hateful, davidson2017automated, founta2018large} and libelous identifiers of race, gender, religion, and culture \cite{Silva_Mondal_Correa_Benevenuto_Weber_2021}. This presents a rather myopic approach, focusing on slur words without considering the text's semantics \cite{schmidt-wiegand-2017-survey, rottger-etal-2021-hatecheck}. In real-world discourse, hate speech stems from extremist views and prejudiced perspectives posted in response to a real-world event. Besides, hate speech's dynamics, syntax, and semantics are bound to change following new triggers \cite{gao-etal-2017-recognizing, florio2020time}. \emph{Thus, identifying hateful posts in such scenarios requires going beyond language and keyword-based reliance and comprehending the post's topical and contextual information.} This makes hate speech detection highly contextual, necessitating topical knowledge \cite{masud2021hate}, commonsense reasoning \cite{alkhamissi2022token}, comprehension of stereotypes \cite{warner-hirschberg-2012-detecting}, and cultural references \cite{sue2010microaggressions, breitfeller-etal-2019-finding}.

\begin{figure}[!t]
    \centering
    \includegraphics[width=\columnwidth]{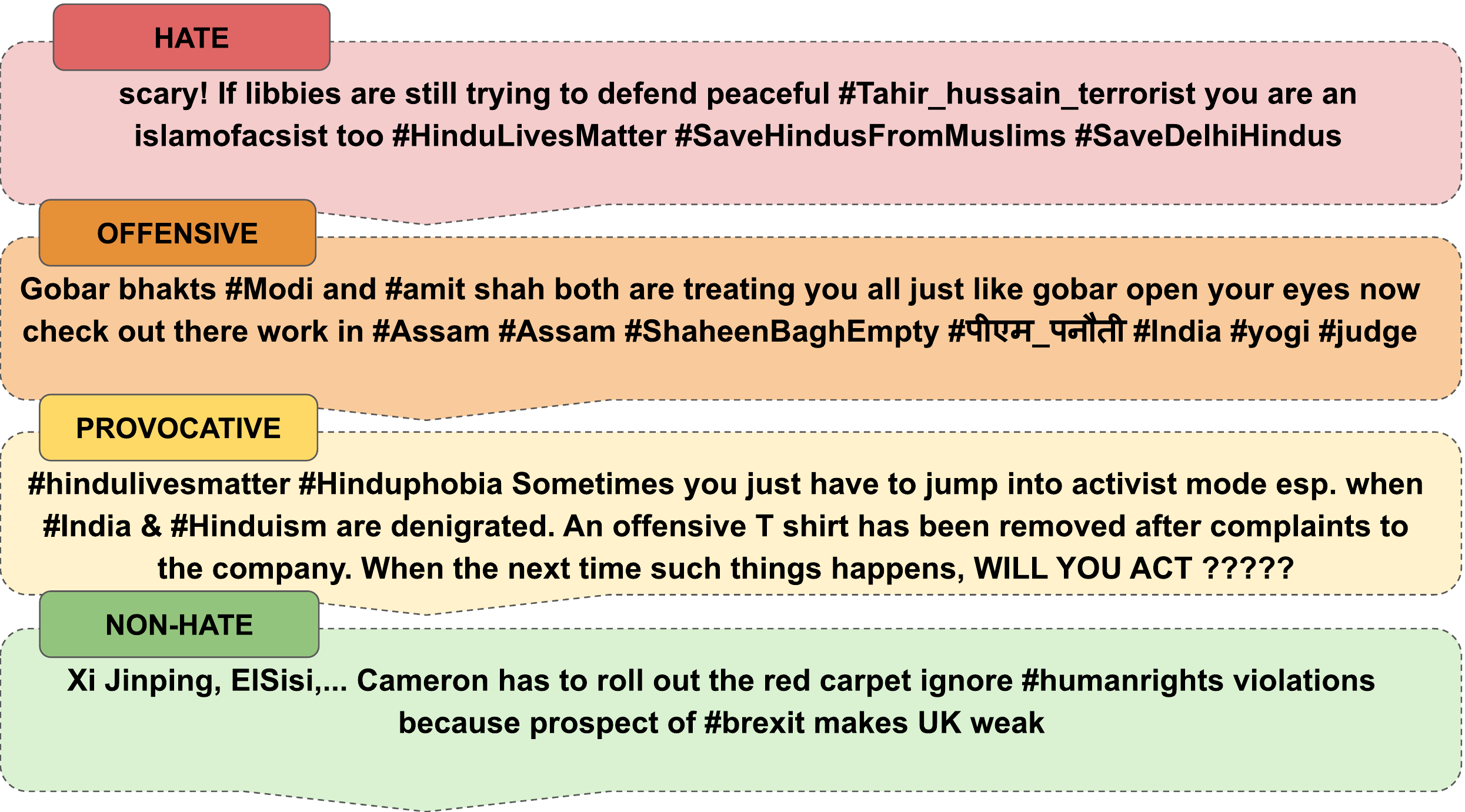}
        \caption{Examples (verbatim) of hateful, offensive, provocative, and non-hateful tweets from \dataset.}
    \label{fig:hopn_eg}
    \vspace{-5mm}
\end{figure}

Moreover, the lexicon strategy yields highly biased data, reflecting annotations far too skewed to accurately represent the real-world distribution of hate speech online \cite{gao-etal-2017-recognizing}. It also leads to low coverage of nuanced hate speech constructs \cite{elsherief-etal-2021-latent}. Furthermore, non-obvious hate speech reduces the semantic diversity between hateful and benign posts \cite{info13060273}. The issue is exacerbated in the case of fine-grained hate speech classification, wherein the already indistinct label boundaries \cite{info13060273} get even blurrier. While a few neutrally-seeded datasets also exist \cite{de-gibert-etal-2018-hate, basile-etal-2019-semeval}, they tend to focus more on controversial events (e.g., the Lee Rigby murder) or specific hate targets (e.g., immigrants), which may introduce topic bias and artificially inflate model performance. Therefore, a dataset covering diverse but disparate topics and languages is the need of the hour.

{\textbf{Proposed dataset:}} To this end, we curate \datasetfull\ (\dataset), a comprehensive fine-grained hate speech classification dataset encompassing nuanced and realistic aspects of online hateful discourse. \emph{\dataset\ is a conglomerate of tweets from seven assorted topics spanning socio-political events and world affairs, each annotated with one of the four labels of hate, offensive, provocative, and non-hate.} We follow a neutral seeding policy, wherein the data is collected for each topic across relevant periods and then annotated for fine-grained hate labels. Therefore, \emph{the dataset is characterized by more cohesion within the topics, resulting in less linguistic, syntactic, and contextual diversity across the labels.} Besides, the dataset covers English, Hindi, and Hinglish, exhibiting linguistic variations. In summary, \dataset\ is a competitive dataset miming real-world online hate speech discourse.  \emph{Note the dataset is not touted as a general reference dataset since such a concept is difficult to define for hate speech. We rather obtain a more `generic view' of social media discourse using neutral topic-based scrapping.}

Figure~\ref{fig:hopn_eg} illustrates representative samples of \dataset. The hateful example critically targets the liberals and the Muslim community. However, the tweet's indirect language and requirement of knowledge about `Islamophobia' makes it difficult to mark it as hate. The offensive statement targetting `Narendra Modi' and `Amit Shah' makes a demeaning comparison between the `Gobar' (cow dung) and `Gobar Bhakts' (dung followers) without using any explicit swearwords. The provocative example invokes a call to action in the context of \#HinduLivesMatter. Lastly, the benign example hints at a subtle mockery of `Xi Jinping' but is not harsh enough to mark it as either hateful, offensive, or provocative. In short, \dataset\ can be challenging for hate speech classification models.

{\textbf{Benchmarking methods:}} We benchmark \dataset\ against $10$ baseline models. Further, to capture the intricacies of \dataset, we present a novel model, \modelfull\ (\model). It is a mixture-of-experts variant of multilingual BERT augmented with endogenous signals. The {\em Timeline Module} tactfully subsumes the users' historical data to accentuate any hateful bias present in the user's posting history. The {\em Exemplar Module} retrieves exemplary tweets that empower the model to distinguish similar and dissimilar features between the labels. The {\em Graph Module} extracts information from the users' social connections via its ego network. Finally, the {\em Reasoning Module} aggregates these multiple experts via attention-based mutual interaction. Empirical results attest that \model\ outperforms the best baseline by $2.5\%$ points in macro-F1 and $5\%$ points in hate class F1. Overall, our experiments attest that \dataset\ is a tough dataset for hate speech classification. 

In summary, our main contributions are as follows:
\begin{itemize}[noitemsep,nolistsep,topsep=0pt,leftmargin=1em]
    \item \textbf{Novel dataset:} We release a neutrally seeded, hate speech dataset that spans seven topics, three geographies and three languages. Additionally, we provide detailed guidelines for annotation and introduce a new category of provocation (Section \ref{sec:dataset}). 
    \item \textbf{Cross-dataset study:} We perform various experiments to analyze \dataset\ against the existing hate speech datasets and establish the challenges \dataset\ brings in  (Section \ref{sec:yet_another}).
    \item \textbf{Incorporation of endogenous signal:} We also experiment with various endogenous signals to enhance the detection of hateful content. The combination of such extensive primary and auxiliary data, combining limited labeled information with plentiful unlabeled but contextual information, nudges for a thorough and systematic study of hate speech detection (Section \ref{sec:method}). 
    \item \textbf{Benchmarking:} We benchmark \dataset\ with ten diverse and widely-studied baseline methods (Section \ref{sec:exp}). 
    \item \textbf{Content moderation pipeline:} This research has led to the creation of a hate speech detection pipeline currently under development in collaboration with Wipro AI \cite{sarahkdd22,masud2021hate} (Section \ref{sec:deployment}).
\end{itemize}

\noindent{\textbf{Reproducibility:}} The source code and sample dataset are publicly available on our Github\footnote{\url{https://github.com/LCS2-IIITD/GotHate}}. Given the data-sharing policy of Twitter, we will not be able to release the tweet text publicly, but we will realize the full set of tweet id and their labels. Additionally, the full dataset (in text and network form) will be available to researchers upon request, as is the case with the existing hate speech data-sharing policy. Limitations and Ethical considerations regarding our dataset and annotation are outlined in Appendices \ref{app:lim} and \ref{app:ethical}. The system can also be extended to other online forums with platform-specific endogenous features, such as LinkedIn, Reddit, etc.

\section{Motivation}
\label{sec:motivation}
\emph{\textbf{Hypothesis I:} There are extremely few semantic and linguistic differences in real-world online hate speech rhetoric around a given topic. Moreover, there is a dearth of dataset that does not accentuate the concept of hatred and represents authentic, real-world speech.} Thus, akin to real-world, \dataset\ features users who post hateful and non-hateful tweets. This intersection, coupled with the topical diversity, neutral seeding, and multi-linguality, makes it a bona fide hate speech corpus that is more difficult to classify than its counterparts. We empirically establish the same in Section \ref{sec:yet_another}.

\emph{\textbf{Hypothesis II:} Most contemporary hate speech datasets are purely text-based with no metadata information, which could be crucial for accurate classification.} Thus, in \dataset, we add endogenous signals (auxiliary data) and combine them with the incoming posts (primary data). We capture a user's intrinsic behavior by incorporating posting history. We consider each  user's ego network to encompass which other users influence their opinions. One can also employ the similarity and diversity from the same label set present within \dataset. The use of exemplar samples can imbue this. The intuition for the individual endogenous signals is highlighted in Section \ref{sec:method}. The advantage of including each signal is discussed during performance comparison and error analysis in Section \ref{sec:exp}.

\emph{\textbf{Hypothesis III:} The content moderators in the real world do not operate in isolation but rather consider the contextual information of who is posting what. By incorporating endogenous signals for hate detection, we aim to replicate the meta-data-rich dashboards that content moderators can access when moderating for a social-media platform.} Therefore, we aim to bring manual and automated content moderation closer. Our experiments further showcase a gap in automated speech detection systems, which can best operate with a human-in-the-loop (Section \ref{sec:deployment}). 
% To better facilitate this, we are partnering with Wipro AI India to develop a semi-automated content moderation and hate speech detection pipeline as outlined in Section \ref{sec:deployment}. 

\section{Dataset}
\label{sec:dataset}
For this study, we concentrate on hateful content on Twitter, owing to the ease of extracting publicly-available posts and their metadata. 

{\textbf{Primary data:}} \dataset\  consists of tweets classified into four categories -- Hate (H), Offensive (O), Provocative (P), and Neutral/non-hate (N). The examples span English, code-mixed Hinglish, and Devanagari Hindi. Using the Twitter API\footnote{\url{https://developer.twitter.com/}}, we compile a corpus of $51,367$ tweets posted by $25,796$ unique users, henceforth referred to as root tweets and root users, respectively. Table \ref{tab:data_stat} presents the topical and label-based overview of \dataset. It spans seven socio-political events (topics) across $3$ geographies (USA, UK, and India). Instead of relying on hate lexicons, we collect the tweets using neutral topics, such as {\em `demonetization' (India)} and {\em `Brexit' (UK)} that garner varying degrees of hate, backlash, and support.
\begin{figure*}[!t]
    \centering
    \subfloat[]{\includegraphics[width=\columnwidth]{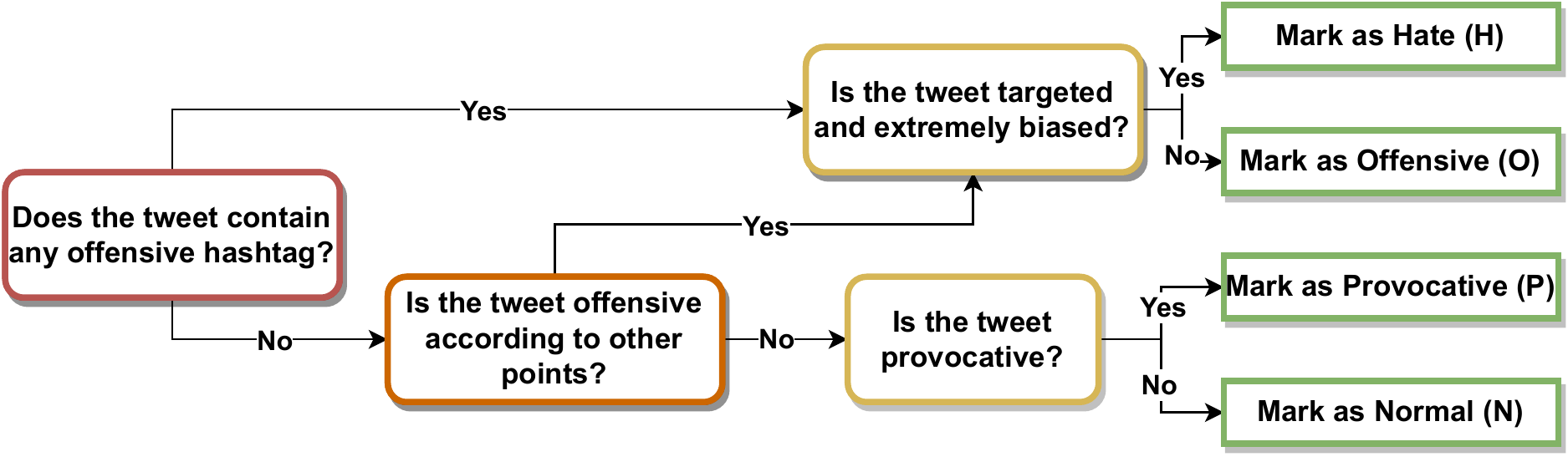}}\hspace{1.5mm}
    \subfloat[]{\includegraphics[width=\columnwidth]{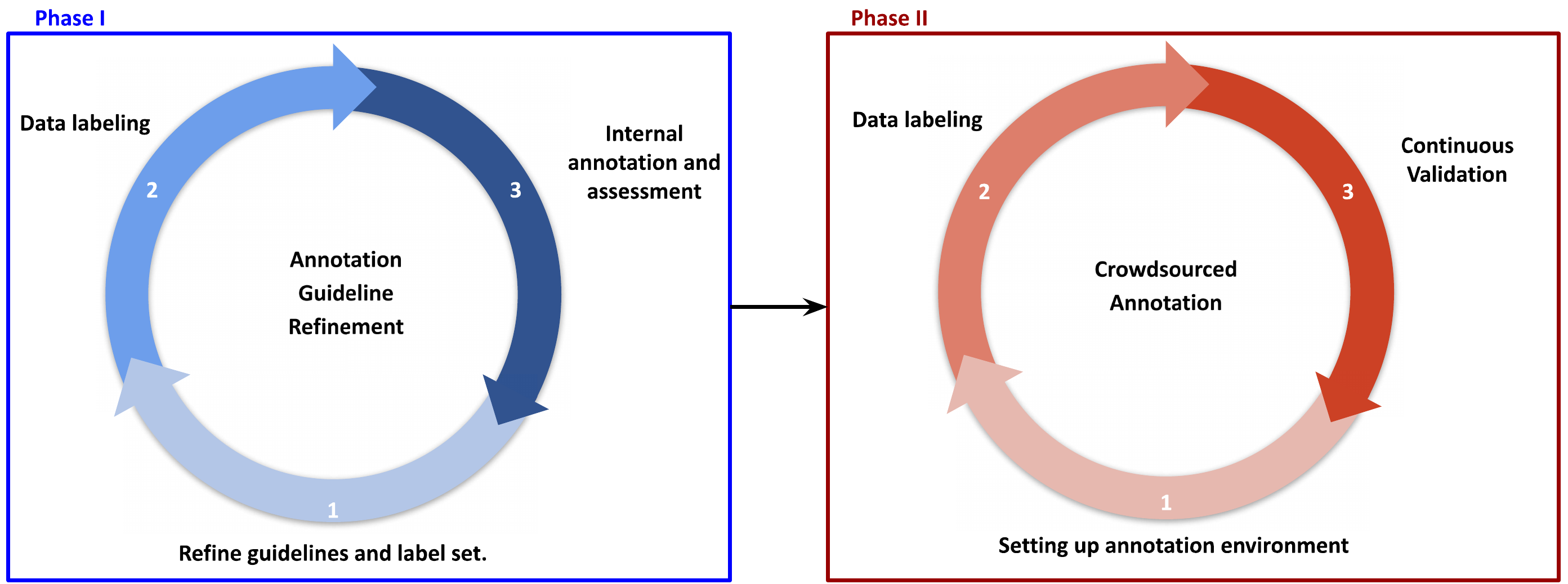}}
    \caption{(a) Flowchart for annotating a post. Precedence for labels: H>O>P>N. (b) Overview of the two-phased continuous-validation annotation process. The label set consists of Hate (H), Offensive (O), Provocative (P) and Neutral (N).}
    \label{fig:annotation}
    \vspace{-2mm}
\end{figure*}
\begin{table}[!h]
\centering
\caption{Statistical information of \dataset\ with unique tweets (users) per label. The user participating in a topic/label subset may not be exclusive to that subset. Label set consists of Hate (H), Offensive (O), Provocative (P) and Neutral (N).}
\label{tab:data_stat}
\resizebox{\columnwidth}{!}{
\begin{tabular}{|p{9em}|c|c|c|c|}
\hline
\textbf{Topic} & \multicolumn{4}{c|}{\textbf{Label-wise unique tweets (users)}} \\\cdashline{2-5}
\textbf{(country of origin)}  & \textbf{H} & \textbf{O} & \textbf{P} & \textbf{N}\\ \hline
Never Trump Campaign (USA) & $5$ ($5$) & $273$ ($222$) & $536$ ($334$) & $1443$ ($889$) \\\hline
Delhi Roits 2020 (India) & $1503$ ($1142$) & $2071$ ($1526$) & $4776$ ($3248$) & $8508$ ($4795$) \\\hline
Demonetization (India) & $574$ ($495$) & $915$ ($750$) & $1446$ ($1154$) & $3643$ ($2112$) \\ \hline
Brexit (UK) & $38$ ($35$) & $505$ ($423$) & $1041$ ($805$) & $6543$ ($3654$) \\ \hline
Umar Khalid JNU (India) & $70$ ($63$) & $2465$ ($1967$) & $18$ ($17$) & $2922$ ($2028$) \\ \hline
Northeast Delhi Riots 2020 (India) & $1182$ ($1055$) & $1622$ ($1441$) &
$1605$ ($1470$) & $4364$ ($3539$) \\\hline
Hindu Lives Matter (USA \& India) & $343$ ($216$) & $219$ ($167$) & $1134$ ($552$) & $1603$ ($967$) \\\hdashline
\textbf{OVERALL} & $51367$ & $8070$ ($6161$) & $10556$ ($7088$) & $29026$ ($16186$)\\\hline 
\end{tabular}
}
\vspace{-5mm}
\end{table}

Interestingly, we find overlapping posts between the topics of {\em `never Trump'} and the Indian protests of Citizenship Amendment Act\footnote{\url{bit.ly/3wWvDRb}}. This can be attributed to Donald Trump's visiting India during the week when protests against the CAA-NRC were at their peak. Our data collection reflects the natural discourse on social media more closely as hateful and provocative content is interlaced with neutral commentary on a topic. The same user can invoke hateful and non-hateful sentiments depending on the topic under discussion \cite{masud2021hate}. The use of hate lexicons may not capture the same. 

{\textbf{Auxiliary data:}} We also collect the root users' timelines and follower networks. The root user's timeline data is curated as the $25$ tweets before and after the posting of the root tweet. We collect the one-hop followers and followees of the root users. Overall, the interaction network consists of $\approx20M$ unique users. We also collect the last $100$ retweeters of every root tweet. Some of these retweeters are followers of the root users, and are therefore marked as internal retweeters. Retweeters not initially captured in our interaction network are marked as external retweeters.

\subsection{Label Definitions}
\emph{\textbf{Vulnerable groups} can be defined as people who have historically been oppressed or abused based on religion, caste, country, color, gender, ethnicity, etc.} For example, Muslims are a vulnerable group in China, whereas, in the USA, people of Chinese origin are minorities. Similarly, religion and caste are some known prejudices within the Indian subcontinent. Therefore, based on the content of a post and the vulnerable group it targets, a post can be considered Hateful (H), Offensive (O), Provocative (P), or Normal (N). As shown in Figure \ref{fig:annotation}(a), to further reduce the disagreement, we introduce an order of priority $H>O>P>N$. While anyone can offend and provoke anyone, hate only applies if the attack is against a vulnerable group. Note that we do not have a pre-defined list of vulnerable groups, but use the definition to help the annotators be cognizant of the task at hand. We define each label as follows: 

\emph{\textbf{Hate:}} It is differentiated by extreme bias\footnote{UN Definition: \url{bit.ly/3HoFpjP}} against the target \cite{waseem-hovy-2016-hateful} via any of the following:
\begin{enumerate}[noitemsep,topsep=0pt,nolistsep, leftmargin=1.5em]
    \item Negatively stereotypes or distorts views on a vulnerable group with unfounded claims.
    \item Silence or suppress a member(s) of a vulnerable group.
    \item Promotes violence against a vulnerable group member(s).
\end{enumerate}

\emph{\textbf{Offensive:}} A statement is offensive if it conforms to one of the following points:
\begin{enumerate}[noitemsep,topsep=0pt,nolistsep, leftmargin=1.5em]
    \item Use of derogatory words to abuse, curse (``go die," ``kill yourself"), sexualize (``f** you," ``kiss my a**"), or express inferiority (``useless," ``dumb," ``crazy") towards an entity, criticizing the person and not the action/event.
    \item Comparison with demons/criminals/animals either directly or by implication (``Wasn't X bad enough," ``Y is p**"). 
    \item Use hashtags covered by either points $\#1$ or $\#2$. Hashtags like \#EUnuch (UK) \#NastyNancy (USA), \#SellOutHiliary (USA), \#PMNautanki (India), \#CoronaJihad (India) are hurtful by themselves. Meanwhile, \#DelhiViolence or \#NeverTrump or  \#NotMyPM, or \#Resign are not offensive just by themselves and need the content of a tweet to determine their label.
\end{enumerate}

\emph{\textbf{Provocative:}} If a post itself is not offensive based on the above definitions but provokes some form of negative reaction from the reader, then it is provocative based on any/all of the following: 
\begin{enumerate}[noitemsep,topsep=0pt,nolistsep, leftmargin=1.5em]
    \item It accuses a particular group or individual of an event related to issues surrounding the group.
    \item Invokes a call to action to stir a group against the target.
    \item The implication of boycotting the target because of issues surrounding the group is also provocative. It can be social (like disallowing entry), economic (like denying financial entitlements), or political (like denying political participation).
\end{enumerate}

\begin{table}[!t]
% \vspace{-3mm}
\centering
\caption{Inter-class similarity within a hate dataset measured via Jensen–Shannon divergence (JS). The lower the JS, the harder it will be to separate the classes.}
\label{tab:js_all_datasets}
\resizebox{\columnwidth}{!}{
\begin{tabular}{|c|c|c|c||c|c|c|c|c|}
\hline
\textbf{Dataset} & \multicolumn{2}{c|}{\textbf{Label}} & \textbf{JS} & \textbf{Dataset} & \multicolumn{2}{c|}{\textbf{Label}} & \textbf{JS}  \\\hline
\multirow{6}{*}{\textbf{\dataset}}   & \textbf{H} & \textbf{O} & \textbf{0.135} &     \multirow{6}{*}{Founta \cite{founta2018large}} & H & A & 0.205 \\
                            & \textbf{H} & \textbf{P} & \textbf{0.087} &                             & H & S & 0.476 \\
                            & \textbf{H} & \textbf{N} & \textbf{0.118} &                             & H & N & 0.319 \\
                            & O & P & 0.119 &                             & A & S & 0.468 \\
                            & O & N & 0.121 &                             & A & N & 0.318 \\
                            & \textbf{P} & \textbf{N} & \textbf{0.063} &                             & S & N & 0.243 \\ \hdashline
\multirow{3}{*}{Davidson \cite{davidson2017automated}}   & H & O & 0.204 & HASOC19 \cite{HASCO19}                     & H & N & 0.246\\  \cdashline{5-8}
                            & H & N & 0.347 & OLID  \cite{zampierietal2019}                        & O & N & 0.115 \\ \cdashline{5-8}
                            & O & N & 0.332 & HatEval \cite{basile-etal-2019-semeval}                    & H & N & 0.131 \\ \hdashline
ImpHate \cite{elsherief-etal-2021-latent}                     & H & N & 0.122 & RETINA \cite{masud2021hate}                       & H & N  & 0.254 \\ \hline
\end{tabular}
}
\vspace{-5mm}
\end{table} 
\subsection{Data Annotation}
We perform annotation in a two-phase continuous validation manner \cite{founta2018large} as shown in Figure \ref{fig:annotation}(b). 

\noindent{\textbf{Annotation phase I:}}
Two  researchers and a lexicographer conducted the first annotation phase (referred to as \textit{Group A}). They were given a random sample of $1000$ tweets spread across all topics. The guidelines crystallized over iterations. During annotation, \textit{Group A} observed that some content had provocative connotations while being neutral as per the Twitter guidelines. We thus introduced `provocation' as a fourth category. Annotation guidelines were refined until Krippendorf's $\alpha$ \cite{Krippendorff2011ComputingKA} of $0.8$ was achieved. 

\noindent{\textbf{Annotation phase II:}}
We partnered with a professional data annotation company for crowdsourced annotation of the entire dataset. %Xsaras\footnote{https://xsaras.com/}. 
After the initial screening, we were assigned a group of $10$ professional annotators (referred to as \textit{Group B}). The corpus was divided into batches of $2500$ samples. The annotation agreement was calculated for $100$ random tweets per batch (annotated by \textit{Group A}) to ensure the annotation quality in each batch. We obtain an average agreement of $0.71$ Krippendorf's $\alpha$. Appendix \ref{app:anno_pipe} provides details about the annotator's demographics. 

{\textbf{Observations from data annotation:}}
Data collection and annotation exercises show us that -- (a) Compared to an annotator agreement of $0.80$ in phase I, we obtain an agreement of $0.71$ at the end of phase II. This reinforces the difficulty in annotating hate speech \cite{Poletto2020, schmidt-wiegand-2017-survey} encompassing the diversity across languages, geographies, and topics. (b) $60$\% of disagreements were in the provocative class. The difference stems from the subtle nature of provocation. Sample annotations are enlisted in Appendix \ref{app:anno_def}.

\section{Yet Another Hate Speech Dataset?}
\label{sec:yet_another}
Given the cornucopia of hate speech datasets, a comparison of \dataset\ and the existing benchmark datasets is imperative. This section discusses our hypothesis that \dataset\ is hard to classify.

Apart from Davidson \cite{davidson2017automated} and Founta \cite{founta2018large}, we also employ three hate shared tasks -- HASOC19 \cite{HASCO19}, OLID \cite{zampierietal2019}, and HatEval \cite{basile-etal-2019-semeval}. HatEval is a neutrally-seeded dataset. We also compare \dataset\ with the implicit hate corpus \cite{elsherief-etal-2021-latent} (referred to as ImpHate, henceforth). Regarding Indian topical diversity, the RETINA dataset proposed by \cite{masud2021hate} comes closest. An outline of various benchmark datasets is provided in Appendix \ref{app:benchmark_datasets}. Meanwhile, extended experiments from the section are also provided in Appendix \ref{app:yet_mbert}. Class labels Hate (H), Offensive (O), Provocative (P), Neutral (N), Abuse (A), and Spam (S) cover all labels in our dataset and the benchmark datasets.

\subsection{Inter-class Similarity}
\label{sec:intra_class}
\noindent\textbf{Intuition.} One can employ Jensen–Shannon (JS) divergence \cite{e21050485} to capture the inter-class proximity within a dataset. The lower the divergence, the closer the class distribution, and the harder it would be to classify and distinguish between them.

\textbf{Analysis.} We generate a Laplacian smoothed uni-gram distribution of each class. Table \ref{tab:js_all_datasets} shows that for \dataset, the JS divergence values for the pairs of H-P=$0.087$ and N-P=$0.063$ are lower than other pairs. This low divergence is a cause for the high disagreement the provocative class receives during annotation and the underlying reason for it. \emph{Additionally, due to the lack of hate lexicons during data collection, the hatred class is closer to neutral (H-N=$0.118$) than the offense (H-O=$0.135$).} On the other hand, the offensive class has a slightly higher divergence from other classes (H-O=$0.135$, P-O=$0.119$, and N-O=$0.121$). Posts containing abusive words and hashtags are more likely to be marked as offensive, providing the class with a latent offensive lexicon. We extend the above inter-class comparison to existing hate speech datasets. In terms of hate-neutral distributions, \dataset\ has a lower divergence ($0.118$) than the more explicit counterparts of Davidson ($0.339$) and Founta ($0.314$). Besides, owing to low explicitness, \dataset's hate-neutral divergence is closer to ImpHate ($0.118$ vs. $0.122$). \emph{Overall, it is evident from  Table \ref{tab:js_all_datasets} that -- (a) for all hate speech datasets, the inter-class divergence tends to be low; (b) in our case, the reason for the further lowering of divergence can be attributed to the topical similarity and use of neutral seeding for data collection.}

\begin{table}[!t]
\caption{Performance comparison of hate speech datasets. We report the class-wise and overall (C) macro-F1 and Matthews correlation coefficient (MCC).}
\label{tab:hate_perf}
\resizebox{\columnwidth}{!}{
\begin{tabular}{|c|*{6}{c|}}
\hline
\textbf{Dataset} & \textbf{H} & \textbf{O/A} & \textbf{P/S} & \textbf{N} & \textbf{C (MCC)}\\\hdashline
\dataset & $\mathbf{0.20}$ & $\mathbf{0.40}$ & $0.35$ & $0.62$ & $\mathbf{0.39}$ ($\mathbf{0.218}$) \\ \hdashline
Founta \cite{founta2018large} & $0.26$ & $0.77$ & $0.41$ & $0.77$ & $0.55$ ($0.460$)  \\ \hdashline
Davidson \cite{davidson2017automated} & $0.30$ & $0.90$ &- & $0.78$ & $0.66$ ($0.585$)  \\ \hdashline
HASOC19 \cite{HASCO19} & $0.59$ & - & - & $0.78$ &$0.68$ ($0.378$)  \\ \hdashline
OLID \cite{zampierietal2019} & - & $0.51$ & - & $0.81$ & $0.66$ ($0.316$)  \\ \hdashline
HatEval \cite{basile-etal-2019-semeval} & $0.61$ & - & - & $\mathbf{0.53}$ & $0.57$ ($\mathbf{0.215}$) \\ \hdashline
ImpHate \cite{elsherief-etal-2021-latent} & $0.59$ & -& - & $0.74$ & $0.67$ ($0.331$)  \\ \hdashline
RETINA \cite{masud2021hate} & $0.59$ & - & - &$0.98$&$0.78$ ($0.570$) \\ \hline
\end{tabular}
}
\vspace{-3mm}
\end{table}

\subsection{How Hard is \dataset\ to Classify?}
\noindent\textbf{Intuition.} Given the lower inter-class divergence of our dataset, we hypothesize that \dataset\ will be hard to classify. 

\textbf{Analysis.} We extend upon the $n$-gram TF-IDF-based logistic regression model employed in inter-class divergence study to compare performance under this setup. To account for the varying positive and negative class sizes, we employ the Matthews correlation coefficient (MCC) to establish the lower correlation of our dataset's prediction. The experimental results are reported in Table \ref{tab:hate_perf}. \emph{It can be observed that: (a) Our dataset has the lowest macro-F1 for overall and the respective class. This low performance is further corroborated by the low MCC scores ($0.2180$) that \dataset\ receives against the setup. (b) HatEval, another neutrally seeded dataset, has a low MCC score ($0.2153$). However, given the binary nature of labels, it still performs better than the $4$-way labeled \dataset.}

\subsection{Adversarial Validation} 
\label{sec:data_drift}
\noindent\textbf{Intuition.} Data drift \cite{8496795} measures the change in feature space between two dataset versions. All samples in the old ({\em aka} source) dataset are labeled as $0$ for analysis. Consequently, all samples in the new (aka target) dataset are labeled as $1$. A simple classifier is trained to predict the labels as \{0,1\}. A high performance indicates discriminatory features between the two versions of the dataset. We extend this to compare existing hate speech datasets with \dataset.
\begin{table}[!t]
\centering
\caption{Evaluating adversarial validation using \dataset\ as the target dataset against existing datasets as the source. We report accuracy (ACC), macro-F1 (F1), ROC-AUC, and Matthews correlation coefficient (MCC).}
\label{tab:data_drift}
\resizebox{\columnwidth}{!}{
\begin{tabular}{|c|c|c|c|c|}
\hline
\multirow{2}{*}{\textbf{Source Dataset}} & \multicolumn{4}{c|}{\textbf{Evaluation Metric}} \\\cline{2-5}
& \textbf{Acc} & \textbf{F1} & \textbf{ROC-AUC} & \textbf{MCC} \\ \hline
Founta \cite{founta2018large}   & $0.98$ & $0.97$ & $0.977$  & $0.949$ \\ \hdashline
Davidson \cite{davidson2017automated} & $0.99$ & $0.99$ & $0.987$   & $0.974$ \\\hdashline
HASOC19 \cite{HASCO19}  & $0.94$ & $0.93$ & $\mathbf{0.926}$   & $\mathbf{0.873}$ \\\hdashline
OLID \cite{zampierietal2019}  & $0.98$ & $0.95$ & $0.973$   & $0.910$ \\\hdashline
HatEval \cite{basile-etal-2019-semeval} & $0.99$ & $0.99$ & $0.985$   & $0.970$ \\\hdashline
ImpHate \cite{elsherief-etal-2021-latent}  & $0.98$ & $0.98$ & $0.977$  & $0.953$ \\\hdashline
RETINA \cite{masud2021hate}  & $0.96$ & $0.96$ & $\mathbf{0.962}$   & $\mathbf{0.925}$\\\hline
 \end{tabular}
}
\vspace{-4mm}
\end{table}
\textbf{Analysis.} The training (T) split from the source hate dataset ($X_{src}^T$) is labeled as ($Y_{src}^T=0$). Meanwhile, the training split from \dataset\ ($X_{trg}^T$) is labelled as ($Y_{trg}^T=1$). Similar labeling is followed for testing ($X^E, Y^E$). We employ a $n$-gram (\{1,2,3\}) based TF-IDF logistic regression model to capture the data drift with \dataset\ as target data. It can be observed from Table \ref{tab:data_drift} that the lower ROC-AUC scores w.r.t HASOC19 ($0.926$) and RETINA ($0.962$) indicate relatively higher similarity of \dataset\ to the Hindi-oriented aspects covered by these datasets. \emph{However, none of the dataset comparisons lead to a ROC-AUC <$.5$ or MCC $\approx0$, which would have indicated that the feature space of \dataset\ is indistinguishable from existing datasets. The results are contrary; the scores vary within a narrow range of $0.021$ ROCU-AUC and $0.101$ MCC. The lowest ROC-AUC (MCC) score is $0.926$ ($0.873$), obtained from HASOC19 (a multi-lingual Hindi corpus). This corroborates the variability in feature space captured by \dataset.}
\begin{table}[!h]
\caption{Cross-dataset performance comparison among \dataset, HASOC19 and RETINA. We report ROC-AUC (Matthews correlation coefficient) on binarised label sets.}
\label{tab:cross_perf}
\resizebox{\columnwidth}{!}{
\begin{tabular}{|c|*{3}{c|}}
\hline
\textbf{Train$\downarrow$Test$\rightarrow$} & \dataset & RETINA \cite{masud2021hate} & HASOC19 \cite{HASCO19} \\\hline
\dataset & - & $0.645$ ($0.128$) & $0.610$ ($0.207$) \\ \hdashline
RETINA \cite{masud2021hate} & $0.512$ ($0.044$) & - & $0.518$ ($0.078$)  \\ \hdashline
HASOC19 \cite{HASCO19} & $0.546$ ($0.101$)  & $0.583$ ($0.079$) & - \\\hline 
\end{tabular}
}
\vspace{-5mm}
\end{table}
\subsection{Can \dataset\ generalise better?}
\noindent\textbf{Intuition.} From our data-drift experiments, we observe that in terms of feature space, HASOC, and RETINA (datasets focused on the Indian context) are closest to \dataset. However, both largely depend on hashtags and explicitness to capture hatefulness. While generalisability is an open challenge for hate speech, we hypothesize that topic-driven neutrally seeded datasets like ours should perform better than their lexicon-driven counterparts. In this case, the intuition is that when we train on \dataset\ and test on HASOC/RETINA, the performance drop will be less than if we reverse the setup (i.e., train on HASOC/RETINA) and test on \dataset.

\textbf{Analysis.} We again extend upon the $n$-gram TF-IDF-based logistic regression model and perform cross-dataset testing. As the label spaces for \dataset vary from binary labels of HASOC19 and RETINA, we binarize \dataset\ into hateful (H, O) and non-hateful (P, N) subsets. As observed from the ROC-AUC and MCC scores from Table \ref{tab:cross_perf}, based on common vogue, we see a clear lack of cross-dataset generalisability among the three. When trained on \dataset and tested on HASOC19, we observe a ROC-AUC of $0.645$. Meanwhile, when trained on HASOC19 and tested on \dataset, we observe the ROC-AUC of $0.512$, i.e., a $0.133$ difference in performance in the two setups. Similar results are observed for the \dataset\ and RETINA pair -- a $0.064$ drop of ROC-AUC when trained on RETINA and tested on \dataset\ than vice versa. None of these setups are ideal for fully capturing hateful context. Still, under existing circumstances, our dataset provides better scope for knowledge transfer. The characteristics that make \dataset\ hard to classify may lend to generalisability. It can be an interesting direction for the future. 

 \begin{SCfigure*}[][h]
    \caption{Model architecture of \model. The embedding for an incoming tweet is obtained from mBERT. The exemplar and timeline modules obtain the input embedding in the form of the mBERT \texttt{CLS} token. Meanwhile, the graph module receives input from the corresponding user's ego network. Each module enhances the respective input embedding via cross-modal attention. Ultimately, the reasoning module concatenates the three endogenous signals to obtain a context-rich embedding, which passes through an attentive-feed forward block for classification.}
    \includegraphics[width=0.65\textwidth]{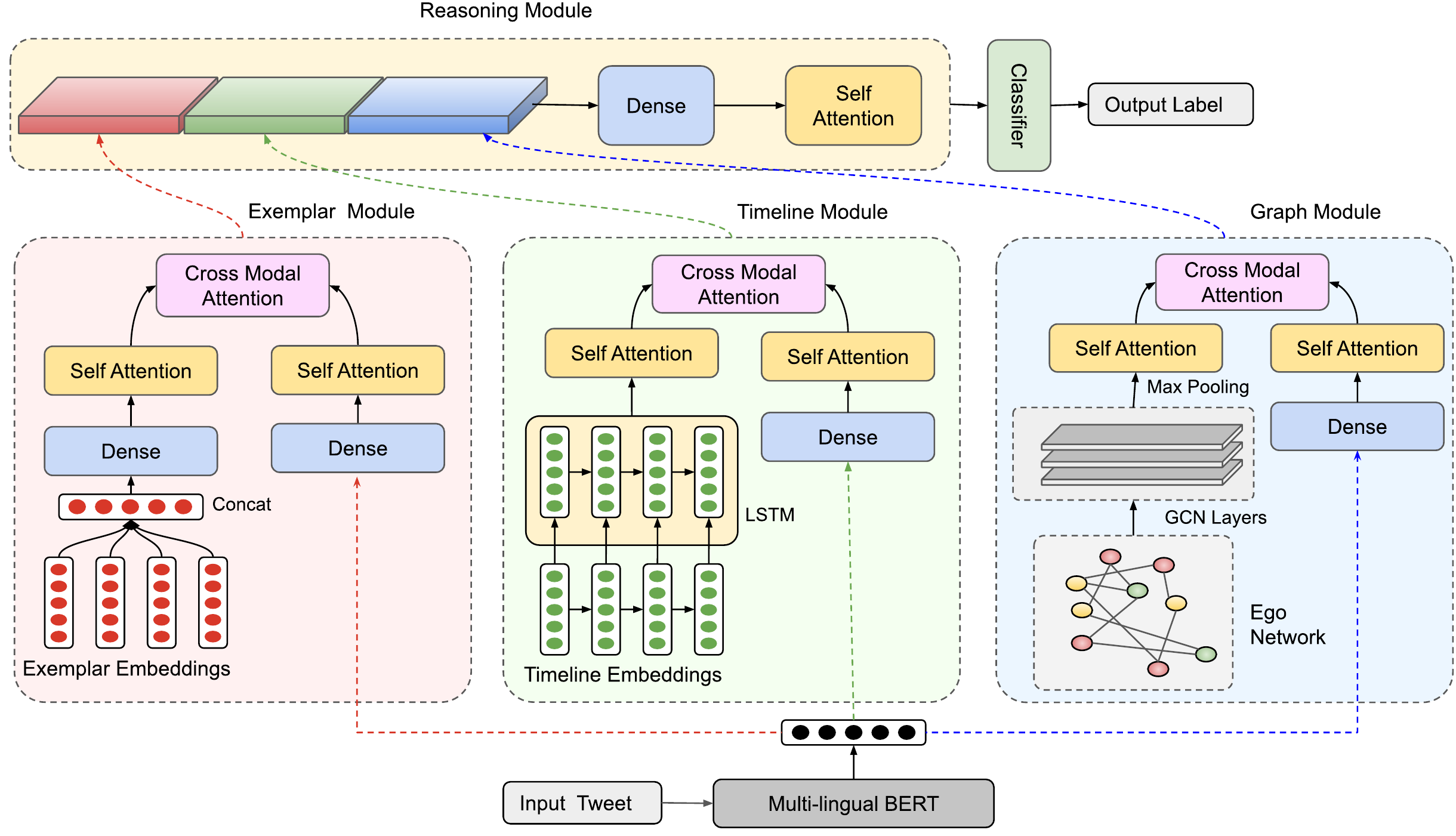}
    \label{fig:model_arch}
\vspace{-2mm}
\end{SCfigure*}
\section{Endogenous Signal Infusion}
\label{sec:method}
This section introduces the modular mixture-of-experts setup called \modelfull\ (\model) that enriches the textual representations with the ancillary signals and improves the detection of hateful content. Figure~\ref{fig:model_arch} illustrates the combined model architecture of \model. 

\textbf{Base Language Module.} We employ multilingual-BERT \cite{devlin-etal-2019-bert} as our base LM and use the \texttt{CLS} token of the last layer to generate embeddings for exemplar and timeline. For the interaction module, we utilize the representations from the last layer of mBERT given by $Z \in \mathbb{R}^{l \times d}$, where $d$ is the feature dimension of LM, and $l$ is the maximum sequence length. 

\subsection{Exemplar Module}
 {\textbf{Intuition.}} Exemplars refer to the set of sample responses from the training set that are semantically related to the input context. They can provide stylistic and thematic cues to the model \cite{schmaltz-2021-detecting}. Such an exemplar-based approach is useful when the inter-label diversity is less, as in the case of \dataset. As one can generate these example samples from within the dataset without scraping extra data, we start by augmenting this signal using dense retrieval via SBERT\footnote{We experimented with BM25 as well, but SBERT performed better.} \cite{reimers-gurevych-2019-sentence}. We employ a label-based exemplar search for the training samples where the exemplars are retrieved from the same label subspace. We extract label invariant (without knowing the label for an incoming test post) exemplars for these validation and test sets using the training dataset for grounding. For each instance, we select the top-$k$ exemplars based on cosine similarity. Some sample exemplars from train and test sets are provided in Appendix \ref{app:exemplar}.

\textbf{Formulation.} To extend and extract this within-dataset signal for an incoming tweet, we concatenate the exemplars $P_i = \{p_{i1},..., p_{ib}\}$ as $F_e \in \mathbb{R}^{b \times d}$. This is followed by a non-linear transformation and dimensionality reduction casting $F_e$ to $\hat F_e \in \mathbb{R}^{b \times d_f}$. The language vector $Z$ also goes through a dimensionality reduction to generate vector $Z_e \in \mathbb{R}^{l \times d_f}$. To extract salient features from $F_e$ and $Z_e$, they are independently passed to a self-attention module \cite{NIPS2017_3f5ee243} to generate $F'_e$ (Eq. \ref{eq:att-feature}) and $Z'_e$ (Eq. \ref{eq:att-lm}), respectively. Finally, these enriched vectors interact with each other via multi-headed cross-modal attention \cite{NIPS2017_3f5ee243} to facilitate deep semantic interaction. Specifically, we query $F'_e$ against $Z'_e$ as the key and the value, generating exemplar-infused language representation $\hat Z_e$ (Eq.~\ref{eq:cross-att-e}).

\begin{minipage}{0.5\linewidth}\scriptsize
\begin{small}
\begin{equation}
  F'_e = Softmax\left(\frac{F_e F_e^T}{\sqrt{d_f}}\right)F_e  \label{eq:att-feature}
\end{equation}
\end{small}
\end{minipage}%
\hfill
\begin{minipage}{0.5\linewidth}\scriptsize
\begin{small}
\begin{equation}
 Z'_e = Softmax\left(\frac{Z_e Z_e^T}{\sqrt{d_f}}\right)Z_e \label{eq:att-lm}
\end{equation}
\end{small}
\end{minipage}
\begin{small}
\begin{equation}
    \hat{Z_e} =  MultiHead\left(Softmax\left(\frac{F_e' Z_e'^T}{\sqrt{d_f}}\right)Z_e'\right) \label{eq:cross-att-e}
\end{equation}
\end{small}
\subsection{Timeline Module}
\noindent{\textbf{Intuition.}} While examplars help obtain latent signals from within the dataset, users' propensity for posting hateful content is not just a one-time incident. Users who interact with similarly malicious users are likely to disseminate hateful content online regularly and more likely to post offensive content than their benign counterparts \cite{10.1145/3415163}. Thus, a user's historical data provides crucial insights into whether they will post hostile material in the future \cite{qian-etal-2018-leveraging}.

\textbf{Formulation.} To begin with, we concatenate historical posts of a user, $T_i = \{t_{i1},..., t_{ia}\}$, to generate feature vector $F_t \in \mathbb{R}^{a \times d}$. To encapsulate the sequential nature of the historical data, $F_t$ undergoes two layers of LSTM to generate temporally-enriched vector $\hat F_t$. Here as well, $Z$ undergoes dimensionality reduction producing vector $Z_t \in \mathbb{R}^{l \times d_f}$. To extract the temporal and language-specific nuances, the vectors $\hat F_t$ and $Z_t$ undergo a self-attention operation, similar to Eq.~\ref{eq:att-feature} and Eq.~\ref{eq:att-lm}. This results in their enriched form of vectors given by $F'_t$ and $Z'_t$. Finally, the language and timeline features interact using multi-headed cross-modal attention. This results in the final feature vector $\hat Z_t$ (Eq. \ref{eq:cross-att-t}) as shown below:
\begin{small}
\begin{eqnarray}
    \hat{Z_t} = MultiHead\left(Softmax\left(\frac{F_t' Z_t'^T}{\sqrt{d_f}}\right)Z_t'\right) \label{eq:cross-att-t}
\end{eqnarray}
\end{small}
\subsection{Graph Module}
\noindent{\textbf{Intuition.}} Hateful users are likelier to follow and retweet other toxic users \cite{10.1145/3415163,ribeiro2018characterizing}. Therefore, we examine their ego network and extract their interaction patterns. We begin by constructing a directed homogeneous network of all the root users, their first-hop followers/followees, and the users retweeting the root users' tweets. We add varying degrees of edge weights to distinguish the different interactions further. Initially, all the edges are given a weight of $1$, except the self-loop weight of $0.1$. Followers/ followees who retweet the root users' tweets earn higher precedence with an edge weight of $1.5$. Meanwhile, external retweeters not included in the root users' follower-followee are given slightly less weight of $0.5$. The base network embeddings are created using  node2vec\footnote{We also experimented with GraphSage \cite{10.5555/3294771.3294869}, but node2vec faired better.} \cite{grover2016node2vec}.

\textbf{Formulation.} For a tweet, $x_i$ by a user $u_j$, \model\ utilizes a user-level ego network $o_j=G(V'_j, E'_j)$ to extract useful interaction patterns. We select the top $100$ followers, followee, and retweeters with the highest node centrality for each user. Each node in $V'_j$ is initialized with node embeddings of dimension $d_g$. The ego network goes through three rounds of graph convolutions \cite{kipf2016semi} followed by a max-pooling operation to spawn an aggregated graph embedding $F_g \in \mathbb{R}^{d_f}$, where $d > d_f > d_g$ (Eq. \ref{eq:gcn}). The language vector $Z$ goes through a dimensionality reduction to generate vector $Z_g \in \mathbb{R}^{l \times d_f}$ (Eq. \ref{eq:transformation}). To extract salient features from $F_g$ and $Z_g$, they are independently passed to a self-attention module \cite{NIPS2017_3f5ee243} to generate $F'_g$ and $Z'_g$, similar to Eq.~\ref{eq:att-feature} and Eq.~\ref{eq:att-lm}. Finally, these enriched vectors interact with each other via multi-headed cross-modal attention. We query $F'_g$ against $Z'_g$ as the key and the value, generating the network-aware language representation $\hat Z_g$ (Eq. \ref{eq:cross-att-g}).
\begin{minipage}{.45\linewidth}\scriptsize
\begin{small}
\begin{equation}
  F_g = GraphConv(G(V'_j,E'_j))  \label{eq:gcn}
\end{equation}
\end{small}
\end{minipage}
\hfill
\begin{minipage}{.45\linewidth}\scriptsize
\begin{small}
\begin{equation}
  Z_g = ReLU(ZW_g + b_g)  \label{eq:transformation}
\end{equation}
\end{small}
\end{minipage}
\begin{small}
\begin{eqnarray}
    \hat{Z_g} = MultiHead\left(Softmax\left(\frac{F_g' Z_g'^T}{\sqrt{d_f}}\right)Z_g'\right) \label{eq:cross-att-g}
\end{eqnarray}
\end{small}
\subsection{Reasoning Module}
The reasoning module aptly coalesces the information from each module via mutual interaction. We concatenate the vectors $\hat{Z_g}$, $\hat{Z_e}$, and $\hat{Z_t}$ and apply a non-linear transformation for dimensionality reduction, resulting in vector $Z_{final} \in  \mathbb{R}^{l \times d}$. This vector is further enhanced using a multi-headed self-attention mechanism. The final vector is then passed through a classification head.

\section{Experiments and Results}
\label{sec:exp}
This section outlines the results and error analysis of \model\ and its variants compared to baseline methods when classifying \dataset\ on a 4-way hate detection task (Table \ref{tab:perf}). We additionally enlist the experimental setup for reproducibility in Appendix \ref{app:exp_setup}.

\subsection{Performance Comparison}
\label{sec:perf_compare}
Due to the lack of auxiliary signals in other hate datasets, we cannot extend \model\ to test those. However, we compare endogenous signal augmentation in \model\ against numerous relevant baselines. We additionally enlist the performance for adding each latent signal independently. Due to skewness in label size, we focus on class-wise and overall F1 scores as our performance metric. Table \ref{tab:perf} shows the model level performance. We also provide a detailed class-wise precision and recall breakdown in Appendix \ref{app:perf}.

\textbf{Traditional models:} We employ Naive Bayes (NB), Logistic Regression (LR), and Support Vector Machines (SVM) with $n$-gram based TF-IDF based features. These models (M1-M3), enlisted in Table \ref{tab:perf}, show that the traditional machine learning baselines yield inferior performances on hate and provocation labels. Upon examination (breakdown provided in Appendix \ref{app:perf}), we observe that those systems are characterized by a high recall for the benign samples and a low recall for the hateful ones, indicating that such models are aggressive in predicting each sample as non-hateful. By augmenting the $n$-gram features with semantic features (pos-tags) and textual meta-data (Vader score, hashtag counter, etc.), the Davidson \cite{davidson2017automated} model (M4) slightly improves the performance for the hate class, compared to vanilla statistical baselines. The skewness of the class labels and over-dependency on word co-occurrence contribute to the downfall of these systems.

\textbf{Neural baselines:} In the next set of modeling, we employ vanilla CNN \cite{kim-2014-convolutional} and LSTM \cite{schmidhuber1997long}  coupled with concatenated Glove embeddings of English and Hindi (EN+HI). We observe that CNN and LSTM neural models (M5-M6) fare better than the traditional ones but still give underwhelming results. We notice about $2-5\%$ gain over traditional methods for the hate and provocative instances in terms of the F1 score. The F1 score for offensive labels more or less remains the same. The Founta \cite{10.1145/3292522.3326028} model (M7) with a self-attention-based RNN method and Glove (EN+HI) also reports low performance, especially for hate classes. Static word vectors and their inability to capture context contribute to their pitfalls.

\begin{table}[!t]
\caption{Performance analysis in terms of macro-F1 of competing models as discussed in Section \ref{sec:perf_compare}.}
\label{tab:perf}
\resizebox{\columnwidth}{!}{
\begin{tabular}{|c|*{5}{c|}}
\hline
\textbf{Model} & \textbf{Hate} & \textbf{Offensive} & \textbf{Provocative} & \textbf{Neutral} & \textbf{Combined} \\ \hline

M1: NB & $0.1802$ & $0.4153$ & $0.3004$ & $0.7267$ &$0.4057$ \\\hdashline

M2: LR & $0.1957$ & $0.4430$ & $0.3271$ & $\mathbf{0.7601}$ &$0.4315$ \\\hdashline

M3: SVC & $0.2382$ & $0.4520$ & $0.3519$ & $0.7382$ & $0.4451$ \\\hdashline

M4: Davidson \cite{davidson2017automated} & $0.2504$ & $0.4070$ & $0.3734$ & $0.6069$ & $0.4094$ \\\hline\hline

M5: CNN & $0.2474$ & $0.4148$ & $0.3716$ & $0.6019$ & $0.4087$ \\\hdashline

M6: LSTM & $0.2579$ & $0.4508$ & $0.4052$ & $0.6352$ & $0.4373$ \\\hdashline

M7: Founta \cite{founta2018large} & $0.2075$ & $0.3603$ & $0.2864$ & $0.5650$ & $0.3548$ \\\hline\hline

M8: mBERT & $0.2795$ & $0.4939$ & $0.3856$ & $0.7285$ & $0.4719$ \\\hdashline

M9: ARHNet \cite{ghosh-chowdhury-etal-2019-arhnet}& $0.2860$ & $0.4719$ & $0.4075$ 
& $0.7140$ & $0.4699$ \\\hdashline

M10: HurtBERT \cite{koufakou-etal-2020-hurtbert} & $0.2717$ 
& $0.4937$ 
& $0.4024$ 
& $0.7143$ 
& $0.4705$ \\\hline\hline

M11: \model\textsubscript{E} & $0.3090$ & $0.5019$ & $\mathbf{0.4395}$ & $0.6977$ & $0.4870$ \\\hdashline

M12: \model\textsubscript{T} & $0.2894$ & $0.4853$ & $0.4271$ & $0.6987$ & $0.4751$ \\\hdashline

M13: \model\textsubscript{G} & $0.2829$ & $\mathbf{0.5189}$ & $0.3891$ & $0.6938$ & $0.4712$ \\\hdashline

M14: \model & $\mathbf{0.3367}$ & $0.4997$ & $0.4061$ & $0.7358$ & $\mathbf{0.4946}$ \\\hline  
\end{tabular}
}
\vspace{-5mm}
\end{table}

\textbf{Transformer baselines:} By applying mBERT (M8), we observe substantial gain over traditional and neural baselines -- a $2\%$, $4\%$, $3\%$, and $4\%$ improvement in F1-score for hate, offensive, provocative, and non-hate labels, respectively. Additionally, we train two existing hate detection models, ARHNet (M9) \cite{ghosh-chowdhury-etal-2019-arhnet} and HurtBERT (M10) \cite{koufakou-etal-2020-hurtbert}. ARHNet concatenates the language representations with node2vec user embeddings as a late fusion combination. For a fairer comparison, we train an mBERT-based ARHNet instead of the original BiLSTM based. Though we see about $1\%$ improvement for the hated class, overall, it does not fare better than naive mBERT. On the other hand, the HurtBERT model, which jointly encodes multilingual hate lexicon knowledge with the language representations, does not showcase improvements over mBERT. This might be attributed to the absence of large-scale hateful lexicons in \dataset\ that impede HurtBERT's performance.

\textbf{Endogenous signal infusion:} The based mBERT that incorporates all three endogenous signals is \model\ (M14). The individual latent augments are enlisted as \model\textsubscript{\{E, T, G\}} to represent the addition of either exemplar, historical, or network-based features, respectively. As Table \ref{tab:perf} shows, the performance for  hate classification is enhanced by adding each module. The exemplar-only module (M11: \model\textsubscript{E}) reports an overall increase in performance, presenting noteworthy gains for the hate and provocative class. As these classes are the most difficult to classify, we conjecture that adding exemplary tweets helps the model unravel the stylistic and semantic nuances across these labels. It is especially true for the provocative class, which witnesses the highest disagreement during annotation. While existing baselines reach F1 of $0.4075$ for provocation detection, \model\textsubscript{E} improves it to $0.4395$. Meanwhile, infusing exemplar signals significantly improves ($2.4$ F1 increase) hate classification. While the timeline module (M112: \model\textsubscript{T}) does provide an improvement over existing baselines, the results do not improve upon the exemplar setup. On the other hand, the graph module (M13: \model\textsubscript{G}) scores the highest points for the offensive class. It suggests that the offensive users' interaction patterns are more distinct than those of the other classes. 

\begin{table*}[!t]
\centering
\caption{Error analysis of mBERT and \model\ analyzing examples of \textit{correct classification, misclassification, and mislabelling}.}
\label{tab:error}
\resizebox{\textwidth}{!}{
\begin{tabular}{|p{45em}|c|c|c|}
\hline
\multicolumn{1}{|c|}{\textbf{Test tweet}} & \textbf{Gold} &\textbf{mBERT} & \textbf{\model}\\ \hline
% & & \textbf{mBERT} & \textbf{\model}\\\hline
\textbf{Example 1:} \$MENTION\$ \$MENTION\$ Arvind Kejriwal's AAP is ISIS OF INDIA. HIS PROPAGANDA WILL DESTROY HINDUS FROM INDIA WITH HELP OF CONGRESS AND LEFTIST. SAVE HINDU SAVE BHARAT. SAVE BRAHMIN , SAVE DALITS, SAVE HINDU OTHER CAST delhi burns delhi riots2020 arrest sonia gandhi delhi voilence & O & O & O\\ \hdashline
\textbf{Example 2:} this must have been a apka tahir offer to jihadis (\small terrorist) - kill kaffirs, loot their property, do what u want with any kaffir (\small non-beliver) female that "ur right hand posseses" ...in short, maal-e-ganimat delhi riots delhi riots & H & P & H\\ \hdashline
\textbf{Example 3:} $MENTION$ It's not the number, it's the \% increase and the doubling time. This administration is just not up to it' They thought br exit would be easy and were wrong, and they are in grave danger of causing much misery through vacillation and poor choices. covid19 \$URL & P & P & O\\ \hline
\end{tabular}
}
\vspace{-3mm}
\end{table*}

Finally, combining all three signals in \model\ (M14) summits the performance across all the metrics -- a significant improvement of about $5\%$ in the hate label F1-score (from best baseline) and a $3$ point improvement over best \model variant. The troublesome provocative class also enjoys a $2$\%  rise in the F1 score. Combining all three variants also keeps the neutral class's performance on the higher side ($0.7358$). Collaboratively, we see a $2$ point increase in the combined macro-F1.  Interestingly, from Table \ref{tab:perf}, we observe that the addition of \model\textsubscript{E} gives the highest improvement for provocative classification ($0.4395$). Meanwhile, the addition of \model\textsubscript{G} improves offensive classification to the highest degree ($0.5189$). Combining all signals shows the most significant improvement in the hate class ($0.3367$) and overall ($0.4946$).

Based on the extensive performance comparison for \dataset\ against benchmark models, we observe that: (a) Our attempts at modeling textual features in different forms (from non-contextual to contextual to endogenous informed) corroborate that hate speech detection is not just a text classification task. (b) \model\ and its variants produce a balanced and improved performance, especially for the harder-to-classify classes of hate and provocation. (c) No one auxiliary signal is fully comprehensive. Combining them helps differentiate the latent space between hateful and non-hateful contexts. (d) The above points reiterate our second hypothesis about the usefulness of auxiliary signals in building better hate classification.

\textbf{Note on the proposed framework:}
We experimented with various modeling techniques and observed that simpler but endogenous signal-rich setups worked better. Cognizant that not all internet forums can access all features, we propose using a pluggable framework where the individual modules can be tailored for the use case. Further, the attention-based infusion brings the latent subspaces of different auxiliary signals closer to the textual subspace. Building upon \model\ provided in this study is the way forward for context-aware hate speech detection.

\subsection{Error Analysis}
\label{sec:error_analysis}
Given that \model\ extends the mBERT architecture, it is critical to investigate and compare the quality of their predictions. While \model\ is enriched with signals to provide additional context, mBERT relies only on the information captured within a tweet's text. For tweets that contain explicit offences such as ``APP is ISIS OF INDIA'' and ``arrest Sonia Gandhi'' (\#1 in Table \ref{tab:error}), both the models effectively pick on these cues and make correct predictions. 

However, in examples that require a post's contextual knowledge (\#2 in Table \ref{tab:error}), mBERT falters. A closer analysis of this tweet's users reveals that they actively post Islamophobic content. Consider the following sample tweets from the user's posting history, collected from the user's timeline posted before ($t_b$) and after ($t_a$) the tweet under examination: ($\mathbf{t_b}$) "\textit{\small look at what Hindus living in mixed-population localities are facing, what $MENTION$ had to face for merely asking his Muslim neighbors not to harass his daughter sexually...and even then if u ask why people don't rent to Muslims, get ur head examined.}" ($\mathbf{t_a}$) "\textit{\small $MENTION$ and $MENTION$ naah...Islamists will never accept Muslim refugees; they will tell the Muslims to create havoc in their home countries and do whatever it takes to convert Dar-ul-Harb into Dar-ul-Islam. Something we should seriously consider doing with Pak Hindus too}". Using such information, our model can develop an understanding of this user's hate propensity to make the correct predictions, while mBERT predicts incorrectly.

While most misclassifications by both models can be attributed to the ambiguities of hate speech, one must also consider mislabeled annotations. As observed from \#3 in Table \ref{tab:error}, though the tweet seems innocuous, it was annotated as provocative and predicted the same by mBERT, while \model\ predicts it as offensive. Such mislabelling indicates that no hate speech pipeline is immune to annotation biases.       

\begin{figure*}[!t]
    \centering
    \includegraphics[width=0.80\textwidth]{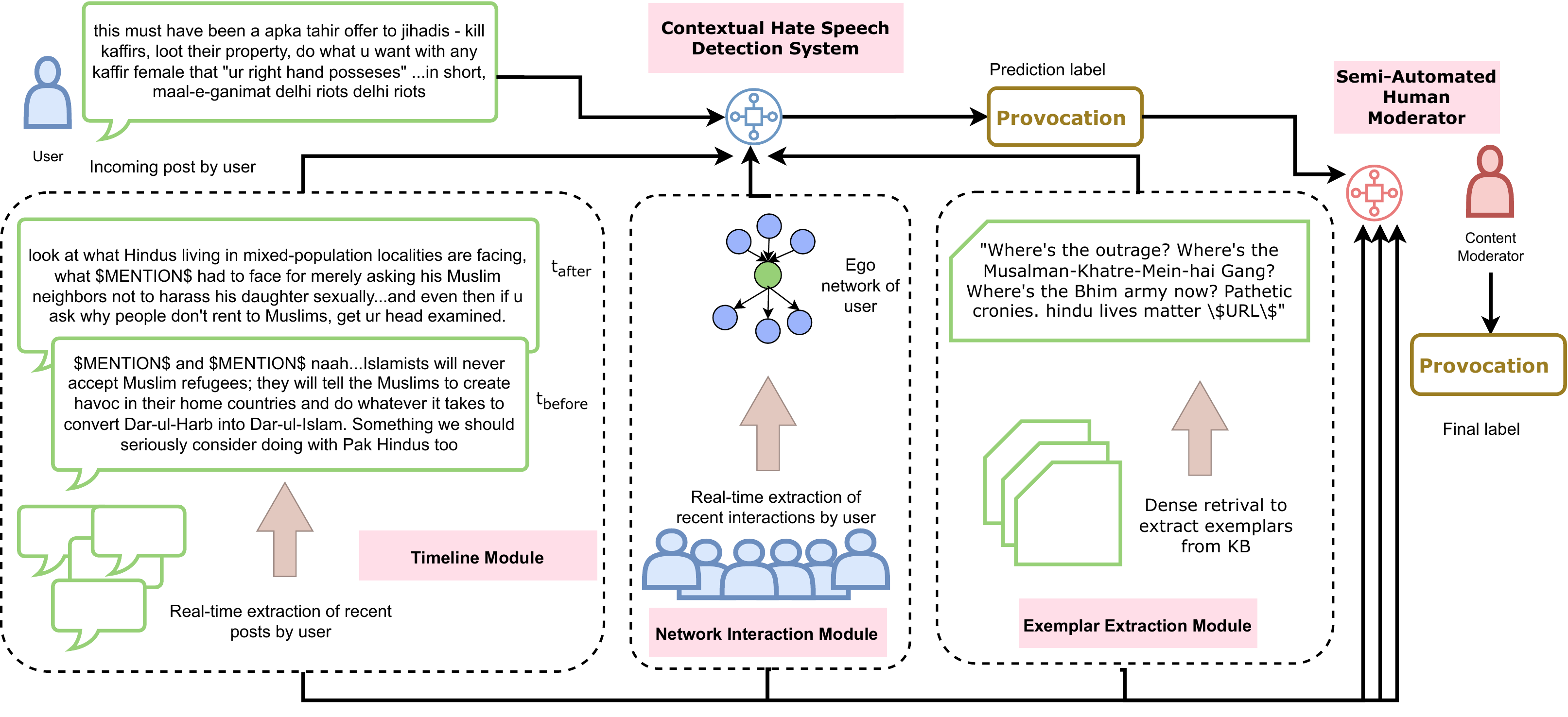}
    \caption{The desired pipeline being developed for semi-automated flagging of hateful content. For an incoming post, we query the user's timeline and user interaction to extract their most recent footprint. Additionally, we query the existing exemplar base to obtain the exemplar. The combination of incoming posts and auxiliary signals is employed for contextual hate speech detection. Meanwhile, these signals and predicted labels will be provided to content moderators to confirm the label.}
    \label{fig:web_deploy}
\vspace{-3mm}
\end{figure*}

\section{Content Moderation Pipeline}
\label{sec:deployment}
Inspired by our experiments, in partnership with Wipro AI, we are developing an interactive web interface for contextual hate speech detection \cite{sarahkdd22}. This interface will be a part of their more extensive pipeline to flag and analyze harmful content on the web. An overview of various components of the semi-automated content flagging interface is outlined in Figure \ref{fig:web_deploy}. In our current experiments, we establish our hypothesis of using endogenous signals by testing in offline mode in which the test set was split for \dataset\, and the auxiliary signals were already curated. However, in our proposed pipeline, the system will operate online. For an incoming tweet posted by a user, we will query their most recent timeline and network interaction in real time. As noted in our experiments, for a slight variation in accuracy, the static node2vec system can be replaced by incremental GraphSage \cite{10.1145/3415163}. Meanwhile, the pre-trained exemplar knowledge base will extract exemplars for an incoming tweet. 
The auxiliary signals thus obtained will be used to perform contextual hate speech detection. The existing input post, its endogenous signals, and the predicted label will aid content moderators in the eventual flagging of content. In future iterations, we aim to use this pipeline to generate feedback from moderators and incrementally train our detection and exemplar models \cite{qian-etal-2021-lifelong}.   

\section{Related Work}
\noindent{\textbf{Hate speech datasets:}} Based on the hypothesis that gender and race are primary targets of hate speech, \citet{waseem-hovy-2016-hateful} released a dataset of $16k$ tweets labeled in-house. \citet{davidson2017automated} released a crowd-sourced annotated dataset of size $25k$ tweets. Later \citet{founta2018large} provided a large-scale corpus of $80k$ English tweets curated via Twitter streaming API. However, they applied bootstrapped sampling to enhance the volume of minority classes. Recently, \citet{toraman2022largescale} proposed another large-scale crowd-sourced dataset for English and Turkish ($100k$ each) curated across five topics. Here too, the authors relied on manually curated keywords to sample tweets per topic. Researchers are now focusing on datasets specific to diverse languages \cite{mulki-etal-2019-l, pavlopoulos-etal-2017-deep, Romim2021, HASCO19, rizwan-etal-2020-hate}. Studies are also being conducted to understand the subtle forms of implicit hate \cite{elsherief-etal-2021-latent, hartvigsen-etal-2022-toxigen}. While a few neutrally-seeded datasets also exist \cite{de-gibert-etal-2018-hate, basile-etal-2019-semeval}, they tend to focus more on a single event or target. 

{\textbf{Methods for hate speech classification:}}
Systems designed to detect hate speech range from employing feature-engineered logistic regression \cite{waseem-hovy-2016-hateful,davidson2017automated} to non-contextual embedding combined with vanilla CNN and LSTM \cite{10.1145/3041021.3054223}. Nowadays, transformer \cite{NIPS2017_3f5ee243} based language models (LM) are being used effectively for hate speech detection \cite{10.1007/978-3-030-36687-2_77,caselli-etal-2021-hatebert}. Methods have also explored using additional tasks to aid hate speech detection, such as emotions \cite{Chiril2022, 10.1007/978-3-030-75762-5_55} and sentiment \cite{10.1145/3394231.3397890, zhou-etal-2021-hate} etc.. Valiant efforts are being made to explore the usage of topical, historical \cite{qian-etal-2018-leveraging}, network data \cite{ghosh-chowdhury-etal-2019-arhnet}.

\section{Conclusion}
The rampant spread of hate speech on social media has serious ramifications for victims across demographics. While characterizing and detecting discriminatory speech online is an active research area, most works have focused on a rather explicit form of hate speech, not accounting for topical heterogeneity and linguistic diversity. In this work, we presented \datasetfull\ (\dataset), a multi-class hate speech dataset that contains minimal slur terms and covers assorted topics across different languages. In summary, \dataset\ manifests as a challenging corpus for classifying hate. We benchmark \dataset\ against several benchmark methods. We further employ \modelfull (\model) -- a mixture of experts that subsumes endogenous knowledge in mBERT to perform contextually rich hate speech detection. Inspired by the utility of endogenous signals, we are collaborating with Wipro AI to develop a feature-rich pipeline for detecting and moderating hateful content. 

\begin{acks}
The authors would like to acknowledge the support of the Prime Minister Doctoral Fellowship (SERB India) and the Wipro Research Grant. We would also like to thank our industry partner Wipro AI. Wipro, an Indian multinational conglomerate with diverse businesses, coordinated the field study for possible deployment. We acknowledge the support of Shivam Sharma, Technical Lead, Wipro AI, for the same. We thank Chhavi Jain and Rituparna Mukherjee for their contributions to data curation. We also thank Xsaras for their support in crowdsourced annotations. 
% We thank all the human subjects for their help in evaluating our tool.
\end{acks}

\newpage
\bibliographystyle{ACM-Reference-Format}
\bibliography{kdd23}


\begin{thebibliography}{65}

%%% the Web address.

\ifx \showCODEN    \undefined \def \showCODEN     #1{\unskip}     \fi
\ifx \showDOI      \undefined \def \showDOI       #1{#1}\fi
\ifx \showISBNx    \undefined \def \showISBNx     #1{\unskip}     \fi
\ifx \showISBNxiii \undefined \def \showISBNxiii  #1{\unskip}     \fi
\ifx \showISSN     \undefined \def \showISSN      #1{\unskip}     \fi
\ifx \showLCCN     \undefined \def \showLCCN      #1{\unskip}     \fi
\ifx \shownote     \undefined \def \shownote      #1{#1}          \fi
\ifx \showarticletitle \undefined \def \showarticletitle #1{#1}   \fi
\ifx \showURL      \undefined \def \showURL       {\relax}        \fi
% invisible to TeX
\providecommand\bibfield[2]{#2}
\providecommand\bibinfo[2]{#2}
\providecommand\natexlab[1]{#1}
\providecommand\showeprint[2][]{arXiv:#2}

\bibitem[AlKhamissi et~al\mbox{.}(2022)]%
        {alkhamissi2022token}
\bibfield{author}{\bibinfo{person}{Badr AlKhamissi}, \bibinfo{person}{Faisal
  Ladhak}, \bibinfo{person}{Srini Iyer}, \bibinfo{person}{Ves Stoyanov},
  \bibinfo{person}{Zornitsa Kozareva}, \bibinfo{person}{Xian Li},
  \bibinfo{person}{Pascale Fung}, \bibinfo{person}{Lambert Mathias},
  \bibinfo{person}{Asli Celikyilmaz}, {and} \bibinfo{person}{Mona Diab}.}
  \bibinfo{year}{2022}\natexlab{}.
\newblock \showarticletitle{ToKen: Task Decomposition and Knowledge Infusion
  for Few-Shot Hate Speech Detection}.
\newblock \bibinfo{journal}{\emph{arXiv preprint arXiv:2205.12495}}
  (\bibinfo{year}{2022}).
\newblock


\bibitem[Alkomah and Ma(2022)]%
        {info13060273}
\bibfield{author}{\bibinfo{person}{Fatimah Alkomah} {and}
  \bibinfo{person}{Xiaogang Ma}.} \bibinfo{year}{2022}\natexlab{}.
\newblock \showarticletitle{A Literature Review of Textual Hate Speech
  Detection Methods and Datasets}.
\newblock \bibinfo{journal}{\emph{Information}} \bibinfo{volume}{13},
  \bibinfo{number}{6} (\bibinfo{year}{2022}).
\newblock
\showISSN{2078-2489}
\urldef\tempurl%
\url{https://doi.org/10.3390/info13060273}
\showDOI{\tempurl}


\bibitem[Awal et~al\mbox{.}(2021)]%
        {10.1007/978-3-030-75762-5_55}
\bibfield{author}{\bibinfo{person}{Md~Rabiul Awal}, \bibinfo{person}{Rui Cao},
  \bibinfo{person}{Roy Ka-Wei Lee}, {and} \bibinfo{person}{Sandra
  Mitrovi{\'{c}}}.} \bibinfo{year}{2021}\natexlab{}.
\newblock \showarticletitle{AngryBERT: Joint Learning Target and Emotion for
  Hate Speech Detection}. In \bibinfo{booktitle}{\emph{Advances in Knowledge
  Discovery and Data Mining}}, \bibfield{editor}{\bibinfo{person}{Kamal
  Karlapalem}, \bibinfo{person}{Hong Cheng}, \bibinfo{person}{Naren
  Ramakrishnan}, \bibinfo{person}{R.~K. Agrawal}, \bibinfo{person}{P.~Krishna
  Reddy}, \bibinfo{person}{Jaideep Srivastava}, {and} \bibinfo{person}{Tanmoy
  Chakraborty}} (Eds.). \bibinfo{publisher}{Springer International Publishing},
  \bibinfo{address}{Cham}, \bibinfo{pages}{701--713}.
\newblock
\showISBNx{978-3-030-75762-5}


\bibitem[Badjatiya et~al\mbox{.}(2017)]%
        {10.1145/3041021.3054223}
\bibfield{author}{\bibinfo{person}{Pinkesh Badjatiya},
  \bibinfo{person}{Shashank Gupta}, \bibinfo{person}{Manish Gupta}, {and}
  \bibinfo{person}{Vasudeva Varma}.} \bibinfo{year}{2017}\natexlab{}.
\newblock \showarticletitle{Deep Learning for Hate Speech Detection in Tweets}.
  In \bibinfo{booktitle}{\emph{Proceedings of the 26th International Conference
  on World Wide Web Companion}} (Perth, Australia) \emph{(\bibinfo{series}{WWW
  '17 Companion})}. \bibinfo{publisher}{International World Wide Web
  Conferences Steering Committee}, \bibinfo{address}{Republic and Canton of
  Geneva, CHE}, \bibinfo{pages}{759–760}.
\newblock
\showISBNx{9781450349147}
\urldef\tempurl%
\url{https://doi.org/10.1145/3041021.3054223}
\showDOI{\tempurl}


\bibitem[Basile et~al\mbox{.}(2019)]%
        {basile-etal-2019-semeval}
\bibfield{author}{\bibinfo{person}{Valerio Basile}, \bibinfo{person}{Cristina
  Bosco}, \bibinfo{person}{Elisabetta Fersini}, \bibinfo{person}{Debora Nozza},
  \bibinfo{person}{Viviana Patti}, \bibinfo{person}{Francisco~Manuel
  Rangel~Pardo}, \bibinfo{person}{Paolo Rosso}, {and} \bibinfo{person}{Manuela
  Sanguinetti}.} \bibinfo{year}{2019}\natexlab{}.
\newblock \showarticletitle{{S}em{E}val-2019 Task 5: Multilingual Detection of
  Hate Speech Against Immigrants and Women in {T}witter}. In
  \bibinfo{booktitle}{\emph{Proceedings of the 13th International Workshop on
  Semantic Evaluation}}. \bibinfo{publisher}{Association for Computational
  Linguistics}, \bibinfo{address}{Minneapolis, Minnesota, USA},
  \bibinfo{pages}{54--63}.
\newblock
\urldef\tempurl%
\url{https://doi.org/10.18653/v1/S19-2007}
\showDOI{\tempurl}


\bibitem[Bilewicz and Soral(2020)]%
        {bilewicz2020hate}
\bibfield{author}{\bibinfo{person}{Micha{\l} Bilewicz} {and}
  \bibinfo{person}{Wiktor Soral}.} \bibinfo{year}{2020}\natexlab{}.
\newblock \showarticletitle{Hate speech epidemic. The dynamic effects of
  derogatory language on intergroup relations and political radicalization}.
\newblock \bibinfo{journal}{\emph{Political Psychology}}  \bibinfo{volume}{41}
  (\bibinfo{year}{2020}), \bibinfo{pages}{3--33}.
\newblock
\urldef\tempurl%
\url{https://doi.org/10.1111/pops.12670}
\showDOI{\tempurl}


\bibitem[Breitfeller et~al\mbox{.}(2019)]%
        {breitfeller-etal-2019-finding}
\bibfield{author}{\bibinfo{person}{Luke Breitfeller}, \bibinfo{person}{Emily
  Ahn}, \bibinfo{person}{David Jurgens}, {and} \bibinfo{person}{Yulia
  Tsvetkov}.} \bibinfo{year}{2019}\natexlab{}.
\newblock \showarticletitle{Finding Microaggressions in the Wild: A Case for
  Locating Elusive Phenomena in Social Media Posts}. In
  \bibinfo{booktitle}{\emph{Proceedings of the 2019 Conference on Empirical
  Methods in Natural Language Processing and the 9th International Joint
  Conference on Natural Language Processing (EMNLP-IJCNLP)}}.
  \bibinfo{publisher}{Association for Computational Linguistics},
  \bibinfo{address}{Hong Kong, China}, \bibinfo{pages}{1664--1674}.
\newblock
\urldef\tempurl%
\url{https://doi.org/10.18653/v1/D19-1176}
\showDOI{\tempurl}


\bibitem[Cao et~al\mbox{.}(2020)]%
        {10.1145/3394231.3397890}
\bibfield{author}{\bibinfo{person}{Rui Cao}, \bibinfo{person}{Roy Ka-Wei Lee},
  {and} \bibinfo{person}{Tuan-Anh Hoang}.} \bibinfo{year}{2020}\natexlab{}.
\newblock \showarticletitle{DeepHate: Hate Speech Detection via Multi-Faceted
  Text Representations}. In \bibinfo{booktitle}{\emph{12th ACM Conference on
  Web Science}} (Southampton, United Kingdom) \emph{(\bibinfo{series}{WebSci
  '20})}. \bibinfo{publisher}{Association for Computing Machinery},
  \bibinfo{address}{New York, NY, USA}, \bibinfo{pages}{11–20}.
\newblock
\showISBNx{9781450379892}
\urldef\tempurl%
\url{https://doi.org/10.1145/3394231.3397890}
\showDOI{\tempurl}


\bibitem[Caselli et~al\mbox{.}(2021)]%
        {caselli-etal-2021-hatebert}
\bibfield{author}{\bibinfo{person}{Tommaso Caselli}, \bibinfo{person}{Valerio
  Basile}, \bibinfo{person}{Jelena Mitrovi{\'c}}, {and}
  \bibinfo{person}{Michael Granitzer}.} \bibinfo{year}{2021}\natexlab{}.
\newblock \showarticletitle{{H}ate{BERT}: Retraining {BERT} for Abusive
  Language Detection in {E}nglish}. In \bibinfo{booktitle}{\emph{Proceedings of
  the 5th Workshop on Online Abuse and Harms (WOAH 2021)}}.
  \bibinfo{publisher}{Association for Computational Linguistics},
  \bibinfo{address}{Online}, \bibinfo{pages}{17--25}.
\newblock
\urldef\tempurl%
\url{https://doi.org/10.18653/v1/2021.woah-1.3}
\showDOI{\tempurl}


\bibitem[Chakraborty and Masud(2022)]%
        {DBLP:journals/corr/abs-2201-00961}
\bibfield{author}{\bibinfo{person}{Tanmoy Chakraborty} {and}
  \bibinfo{person}{Sarah Masud}.} \bibinfo{year}{2022}\natexlab{}.
\newblock \showarticletitle{Nipping in the Bud: Detection, Diffusion and
  Mitigation of Hate Speech on Social Media}.
\newblock \bibinfo{journal}{\emph{CoRR}}  \bibinfo{volume}{abs/2201.00961}
  (\bibinfo{year}{2022}).
\newblock
\showeprint[arXiv]{2201.00961}
\urldef\tempurl%
\url{https://arxiv.org/abs/2201.00961}
\showURL{%
\tempurl}


\bibitem[Chiril et~al\mbox{.}(2022)]%
        {Chiril2022}
\bibfield{author}{\bibinfo{person}{Patricia Chiril},
  \bibinfo{person}{Endang~Wahyu Pamungkas}, \bibinfo{person}{Farah Benamara},
  \bibinfo{person}{V{\'e}ronique Moriceau}, {and} \bibinfo{person}{Viviana
  Patti}.} \bibinfo{year}{2022}\natexlab{}.
\newblock \showarticletitle{Emotionally Informed Hate Speech Detection: A
  Multi-target Perspective}.
\newblock \bibinfo{journal}{\emph{Cognitive Computation}} \bibinfo{volume}{14},
  \bibinfo{number}{1} (\bibinfo{date}{01 Jan} \bibinfo{year}{2022}),
  \bibinfo{pages}{322--352}.
\newblock
\showISSN{1866-9964}
\urldef\tempurl%
\url{https://doi.org/10.1007/s12559-021-09862-5}
\showDOI{\tempurl}


\bibitem[Da~San~Martino et~al\mbox{.}(2019)]%
        {da-san-martino-etal-2019-findings}
\bibfield{author}{\bibinfo{person}{Giovanni Da~San~Martino},
  \bibinfo{person}{Alberto Barr{\'o}n-Cede{\~n}o}, {and}
  \bibinfo{person}{Preslav Nakov}.} \bibinfo{year}{2019}\natexlab{}.
\newblock \showarticletitle{Findings of the {NLP}4{IF}-2019 Shared Task on
  Fine-Grained Propaganda Detection}. In \bibinfo{booktitle}{\emph{Proceedings
  of the Second Workshop on Natural Language Processing for Internet Freedom:
  Censorship, Disinformation, and Propaganda}}. \bibinfo{publisher}{Association
  for Computational Linguistics}, \bibinfo{address}{Hong Kong, China},
  \bibinfo{pages}{162--170}.
\newblock
\urldef\tempurl%
\url{https://doi.org/10.18653/v1/D19-5024}
\showDOI{\tempurl}


\bibitem[Davidson et~al\mbox{.}(2017)]%
        {davidson2017automated}
\bibfield{author}{\bibinfo{person}{Thomas Davidson}, \bibinfo{person}{Dana
  Warmsley}, \bibinfo{person}{Michael Macy}, {and} \bibinfo{person}{Ingmar
  Weber}.} \bibinfo{year}{2017}\natexlab{}.
\newblock \showarticletitle{Automated Hate Speech Detection and the Problem of
  Offensive Language}.
\newblock \bibinfo{journal}{\emph{Proceedings of the International AAAI
  Conference on Web and Social Media}} \bibinfo{volume}{11},
  \bibinfo{number}{1} (\bibinfo{date}{May} \bibinfo{year}{2017}),
  \bibinfo{pages}{512--515}.
\newblock
\urldef\tempurl%
\url{https://ojs.aaai.org/index.php/ICWSM/article/view/14955}
\showURL{%
\tempurl}


\bibitem[de~Gibert et~al\mbox{.}(2018)]%
        {de-gibert-etal-2018-hate}
\bibfield{author}{\bibinfo{person}{Ona de Gibert}, \bibinfo{person}{Naiara
  Perez}, \bibinfo{person}{Aitor Garc{\'\i}a-Pablos}, {and}
  \bibinfo{person}{Montse Cuadros}.} \bibinfo{year}{2018}\natexlab{}.
\newblock \showarticletitle{Hate Speech Dataset from a White Supremacy Forum}.
  In \bibinfo{booktitle}{\emph{Proceedings of the 2nd Workshop on Abusive
  Language Online ({ALW}2)}}. \bibinfo{publisher}{Association for Computational
  Linguistics}, \bibinfo{address}{Brussels, Belgium}, \bibinfo{pages}{11--20}.
\newblock
\urldef\tempurl%
\url{https://doi.org/10.18653/v1/W18-5102}
\showDOI{\tempurl}


\bibitem[Devlin et~al\mbox{.}(2019)]%
        {devlin-etal-2019-bert}
\bibfield{author}{\bibinfo{person}{Jacob Devlin}, \bibinfo{person}{Ming-Wei
  Chang}, \bibinfo{person}{Kenton Lee}, {and} \bibinfo{person}{Kristina
  Toutanova}.} \bibinfo{year}{2019}\natexlab{}.
\newblock \showarticletitle{{BERT}: Pre-training of Deep Bidirectional
  Transformers for Language Understanding}. In
  \bibinfo{booktitle}{\emph{Proceedings of the 2019 Conference of the North
  {A}merican Chapter of the Association for Computational Linguistics: Human
  Language Technologies, Volume 1 (Long and Short Papers)}}.
  \bibinfo{publisher}{Association for Computational Linguistics},
  \bibinfo{address}{Minneapolis, Minnesota}, \bibinfo{pages}{4171--4186}.
\newblock
\urldef\tempurl%
\url{https://doi.org/10.18653/v1/N19-1423}
\showDOI{\tempurl}


\bibitem[ElSherief et~al\mbox{.}(2021)]%
        {elsherief-etal-2021-latent}
\bibfield{author}{\bibinfo{person}{Mai ElSherief}, \bibinfo{person}{Caleb
  Ziems}, \bibinfo{person}{David Muchlinski}, \bibinfo{person}{Vaishnavi
  Anupindi}, \bibinfo{person}{Jordyn Seybolt}, \bibinfo{person}{Munmun
  De~Choudhury}, {and} \bibinfo{person}{Diyi Yang}.}
  \bibinfo{year}{2021}\natexlab{}.
\newblock \showarticletitle{Latent Hatred: A Benchmark for Understanding
  Implicit Hate Speech}. In \bibinfo{booktitle}{\emph{Proceedings of the 2021
  Conference on Empirical Methods in Natural Language Processing}}.
  \bibinfo{publisher}{Association for Computational Linguistics},
  \bibinfo{address}{Online and Punta Cana, Dominican Republic},
  \bibinfo{pages}{345--363}.
\newblock
\urldef\tempurl%
\url{https://doi.org/10.18653/v1/2021.emnlp-main.29}
\showDOI{\tempurl}


\bibitem[Florio et~al\mbox{.}(2020)]%
        {florio2020time}
\bibfield{author}{\bibinfo{person}{Komal Florio}, \bibinfo{person}{Valerio
  Basile}, \bibinfo{person}{Marco Polignano}, \bibinfo{person}{Pierpaolo
  Basile}, {and} \bibinfo{person}{Viviana Patti}.}
  \bibinfo{year}{2020}\natexlab{}.
\newblock \showarticletitle{Time of Your Hate: The Challenge of Time in Hate
  Speech Detection on Social Media}.
\newblock \bibinfo{journal}{\emph{Applied Sciences}} \bibinfo{volume}{10},
  \bibinfo{number}{12} (\bibinfo{year}{2020}).
\newblock
\showISSN{2076-3417}
\urldef\tempurl%
\url{https://doi.org/10.3390/app10124180}
\showDOI{\tempurl}


\bibitem[Fortuna and Nunes(2018)]%
        {fortuna2018survey}
\bibfield{author}{\bibinfo{person}{Paula Fortuna} {and}
  \bibinfo{person}{S\'{e}rgio Nunes}.} \bibinfo{year}{2018}\natexlab{}.
\newblock \showarticletitle{A Survey on Automatic Detection of Hate Speech in
  Text}.
\newblock \bibinfo{journal}{\emph{ACM Comput. Surv.}} \bibinfo{volume}{51},
  \bibinfo{number}{4}, Article \bibinfo{articleno}{85} (\bibinfo{date}{jul}
  \bibinfo{year}{2018}), \bibinfo{numpages}{30}~pages.
\newblock
\showISSN{0360-0300}
\urldef\tempurl%
\url{https://doi.org/10.1145/3232676}
\showDOI{\tempurl}


\bibitem[Founta et~al\mbox{.}(2018)]%
        {founta2018large}
\bibfield{author}{\bibinfo{person}{Antigoni Founta},
  \bibinfo{person}{Constantinos Djouvas}, \bibinfo{person}{Despoina Chatzakou},
  \bibinfo{person}{Ilias Leontiadis}, \bibinfo{person}{Jeremy Blackburn},
  \bibinfo{person}{Gianluca Stringhini}, \bibinfo{person}{Athena Vakali},
  \bibinfo{person}{Michael Sirivianos}, {and} \bibinfo{person}{Nicolas
  Kourtellis}.} \bibinfo{year}{2018}\natexlab{}.
\newblock \showarticletitle{Large Scale Crowdsourcing and Characterization of
  Twitter Abusive Behavior}.
\newblock \bibinfo{journal}{\emph{Proceedings of the International AAAI
  Conference on Web and Social Media}} \bibinfo{volume}{12},
  \bibinfo{number}{1} (\bibinfo{date}{Jun.} \bibinfo{year}{2018}).
\newblock
\urldef\tempurl%
\url{https://ojs.aaai.org/index.php/ICWSM/article/view/14991}
\showURL{%
\tempurl}


\bibitem[Founta et~al\mbox{.}(2019)]%
        {10.1145/3292522.3326028}
\bibfield{author}{\bibinfo{person}{Antigoni~Maria Founta},
  \bibinfo{person}{Despoina Chatzakou}, \bibinfo{person}{Nicolas Kourtellis},
  \bibinfo{person}{Jeremy Blackburn}, \bibinfo{person}{Athena Vakali}, {and}
  \bibinfo{person}{Ilias Leontiadis}.} \bibinfo{year}{2019}\natexlab{}.
\newblock \showarticletitle{A Unified Deep Learning Architecture for Abuse
  Detection}. In \bibinfo{booktitle}{\emph{Proceedings of the 10th ACM
  Conference on Web Science}} (Boston, Massachusetts, USA)
  \emph{(\bibinfo{series}{WebSci '19})}. \bibinfo{publisher}{Association for
  Computing Machinery}, \bibinfo{address}{New York, NY, USA},
  \bibinfo{pages}{105–114}.
\newblock
\showISBNx{9781450362023}
\urldef\tempurl%
\url{https://doi.org/10.1145/3292522.3326028}
\showDOI{\tempurl}


\bibitem[Gao et~al\mbox{.}(2017)]%
        {gao-etal-2017-recognizing}
\bibfield{author}{\bibinfo{person}{Lei Gao}, \bibinfo{person}{Alexis
  Kuppersmith}, {and} \bibinfo{person}{Ruihong Huang}.}
  \bibinfo{year}{2017}\natexlab{}.
\newblock \showarticletitle{Recognizing Explicit and Implicit Hate Speech Using
  a Weakly Supervised Two-path Bootstrapping Approach}. In
  \bibinfo{booktitle}{\emph{Proceedings of the Eighth International Joint
  Conference on Natural Language Processing (Volume 1: Long Papers)}}.
  \bibinfo{publisher}{Asian Federation of Natural Language Processing},
  \bibinfo{address}{Taipei, Taiwan}, \bibinfo{pages}{774--782}.
\newblock
\urldef\tempurl%
\url{https://aclanthology.org/I17-1078}
\showURL{%
\tempurl}


\bibitem[Ghosh~Chowdhury et~al\mbox{.}(2019)]%
        {ghosh-chowdhury-etal-2019-arhnet}
\bibfield{author}{\bibinfo{person}{Arijit Ghosh~Chowdhury},
  \bibinfo{person}{Aniket Didolkar}, \bibinfo{person}{Ramit Sawhney}, {and}
  \bibinfo{person}{Rajiv~Ratn Shah}.} \bibinfo{year}{2019}\natexlab{}.
\newblock \showarticletitle{{ARHN}et - Leveraging Community Interaction for
  Detection of Religious Hate Speech in {A}rabic}. In
  \bibinfo{booktitle}{\emph{Proceedings of the 57th Annual Meeting of the
  Association for Computational Linguistics: Student Research Workshop}}.
  \bibinfo{publisher}{Association for Computational Linguistics},
  \bibinfo{address}{Florence, Italy}, \bibinfo{pages}{273--280}.
\newblock
\urldef\tempurl%
\url{https://doi.org/10.18653/v1/P19-2038}
\showDOI{\tempurl}


\bibitem[Gretton et~al\mbox{.}(2012)]%
        {mmd}
\bibfield{author}{\bibinfo{person}{Arthur Gretton}, \bibinfo{person}{Karsten~M.
  Borgwardt}, \bibinfo{person}{Malte~J. Rasch}, \bibinfo{person}{Bernhard
  Sch{{\"o}}lkopf}, {and} \bibinfo{person}{Alexander Smola}.}
  \bibinfo{year}{2012}\natexlab{}.
\newblock \showarticletitle{A Kernel Two-Sample Test}.
\newblock \bibinfo{journal}{\emph{Journal of Machine Learning Research}}
  \bibinfo{volume}{13}, \bibinfo{number}{25} (\bibinfo{year}{2012}),
  \bibinfo{pages}{723--773}.
\newblock
\urldef\tempurl%
\url{http://jmlr.org/papers/v13/gretton12a.html}
\showURL{%
\tempurl}


\bibitem[Grover and Leskovec(2016)]%
        {grover2016node2vec}
\bibfield{author}{\bibinfo{person}{Aditya Grover} {and} \bibinfo{person}{Jure
  Leskovec}.} \bibinfo{year}{2016}\natexlab{}.
\newblock \showarticletitle{Node2vec: Scalable Feature Learning for Networks}.
  In \bibinfo{booktitle}{\emph{Proceedings of the 22nd ACM SIGKDD International
  Conference on Knowledge Discovery and Data Mining}} (San Francisco,
  California, USA) \emph{(\bibinfo{series}{KDD '16})}.
  \bibinfo{publisher}{Association for Computing Machinery},
  \bibinfo{address}{New York, NY, USA}, \bibinfo{pages}{855–864}.
\newblock
\showISBNx{9781450342322}
\urldef\tempurl%
\url{https://doi.org/10.1145/2939672.2939754}
\showDOI{\tempurl}


\bibitem[Hamilton et~al\mbox{.}(2017)]%
        {10.5555/3294771.3294869}
\bibfield{author}{\bibinfo{person}{William~L. Hamilton}, \bibinfo{person}{Rex
  Ying}, {and} \bibinfo{person}{Jure Leskovec}.}
  \bibinfo{year}{2017}\natexlab{}.
\newblock \showarticletitle{Inductive Representation Learning on Large Graphs}.
  In \bibinfo{booktitle}{\emph{Proceedings of the 31st International Conference
  on Neural Information Processing Systems}} (Long Beach, California, USA)
  \emph{(\bibinfo{series}{NIPS'17})}. \bibinfo{publisher}{Curran Associates
  Inc.}, \bibinfo{address}{Red Hook, NY, USA}, \bibinfo{pages}{1025–1035}.
\newblock
\showISBNx{9781510860964}


\bibitem[Hartvigsen et~al\mbox{.}(2022)]%
        {hartvigsen-etal-2022-toxigen}
\bibfield{author}{\bibinfo{person}{Thomas Hartvigsen}, \bibinfo{person}{Saadia
  Gabriel}, \bibinfo{person}{Hamid Palangi}, \bibinfo{person}{Maarten Sap},
  \bibinfo{person}{Dipankar Ray}, {and} \bibinfo{person}{Ece Kamar}.}
  \bibinfo{year}{2022}\natexlab{}.
\newblock \showarticletitle{{T}oxi{G}en: A Large-Scale Machine-Generated
  Dataset for Adversarial and Implicit Hate Speech Detection}. In
  \bibinfo{booktitle}{\emph{Proceedings of the 60th Annual Meeting of the
  Association for Computational Linguistics (Volume 1: Long Papers)}}.
  \bibinfo{publisher}{Association for Computational Linguistics},
  \bibinfo{address}{Dublin, Ireland}, \bibinfo{pages}{3309--3326}.
\newblock
\urldef\tempurl%
\url{https://doi.org/10.18653/v1/2022.acl-long.234}
\showDOI{\tempurl}


\bibitem[Kim(2014)]%
        {kim-2014-convolutional}
\bibfield{author}{\bibinfo{person}{Yoon Kim}.} \bibinfo{year}{2014}\natexlab{}.
\newblock \showarticletitle{Convolutional Neural Networks for Sentence
  Classification}. In \bibinfo{booktitle}{\emph{Proceedings of the 2014
  Conference on Empirical Methods in Natural Language Processing ({EMNLP})}}.
  \bibinfo{publisher}{Association for Computational Linguistics},
  \bibinfo{address}{Doha, Qatar}, \bibinfo{pages}{1746--1751}.
\newblock
\urldef\tempurl%
\url{https://doi.org/10.3115/v1/D14-1181}
\showDOI{\tempurl}


\bibitem[Kingma and Ba(2014)]%
        {kingma2014adam}
\bibfield{author}{\bibinfo{person}{Diederik~P Kingma} {and}
  \bibinfo{person}{Jimmy Ba}.} \bibinfo{year}{2014}\natexlab{}.
\newblock \showarticletitle{Adam: A method for stochastic optimization}.
\newblock \bibinfo{journal}{\emph{arXiv preprint arXiv:1412.6980}}
  (\bibinfo{year}{2014}).
\newblock
\urldef\tempurl%
\url{https://arxiv.org/abs/1412.6980}
\showURL{%
\tempurl}


\bibitem[Kipf and Welling(2017)]%
        {kipf2016semi}
\bibfield{author}{\bibinfo{person}{Thomas~N. Kipf} {and} \bibinfo{person}{Max
  Welling}.} \bibinfo{year}{2017}\natexlab{}.
\newblock \showarticletitle{Semi-Supervised Classification with Graph
  Convolutional Networks}. In \bibinfo{booktitle}{\emph{5th International
  Conference on Learning Representations, {ICLR} 2017, Toulon, France, April
  24-26, 2017, Conference Track Proceedings}}.
  \bibinfo{publisher}{OpenReview.net}.
\newblock
\urldef\tempurl%
\url{https://openreview.net/forum?id=SJU4ayYgl}
\showURL{%
\tempurl}


\bibitem[Koufakou et~al\mbox{.}(2020)]%
        {koufakou-etal-2020-hurtbert}
\bibfield{author}{\bibinfo{person}{Anna Koufakou},
  \bibinfo{person}{Endang~Wahyu Pamungkas}, \bibinfo{person}{Valerio Basile},
  {and} \bibinfo{person}{Viviana Patti}.} \bibinfo{year}{2020}\natexlab{}.
\newblock \showarticletitle{{H}urt{BERT}: Incorporating Lexical Features with
  {BERT} for the Detection of Abusive Language}. In
  \bibinfo{booktitle}{\emph{Proceedings of the Fourth Workshop on Online Abuse
  and Harms}}. \bibinfo{publisher}{Association for Computational Linguistics},
  \bibinfo{address}{Online}, \bibinfo{pages}{34--43}.
\newblock
\urldef\tempurl%
\url{https://doi.org/10.18653/v1/2020.alw-1.5}
\showDOI{\tempurl}


\bibitem[Krippendorff(2011)]%
        {Krippendorff2011ComputingKA}
\bibfield{author}{\bibinfo{person}{Klaus Krippendorff}.}
  \bibinfo{year}{2011}\natexlab{}.
\newblock \showarticletitle{Computing Krippendorff's Alpha-Reliability}.
\newblock


\bibitem[Lu et~al\mbox{.}(2019)]%
        {8496795}
\bibfield{author}{\bibinfo{person}{Jie Lu}, \bibinfo{person}{Anjin Liu},
  \bibinfo{person}{Fan Dong}, \bibinfo{person}{Feng Gu}, \bibinfo{person}{João
  Gama}, {and} \bibinfo{person}{Guangquan Zhang}.}
  \bibinfo{year}{2019}\natexlab{}.
\newblock \showarticletitle{Learning under Concept Drift: A Review}.
\newblock \bibinfo{journal}{\emph{IEEE Transactions on Knowledge and Data
  Engineering}} \bibinfo{volume}{31}, \bibinfo{number}{12}
  (\bibinfo{year}{2019}), \bibinfo{pages}{2346--2363}.
\newblock
\urldef\tempurl%
\url{https://doi.org/10.1109/TKDE.2018.2876857}
\showDOI{\tempurl}


\bibitem[Mandl et~al\mbox{.}(2019)]%
        {HASCO19}
\bibfield{author}{\bibinfo{person}{Thomas Mandl}, \bibinfo{person}{Sandip
  Modha}, \bibinfo{person}{Prasenjit Majumder}, \bibinfo{person}{Daksh Patel},
  \bibinfo{person}{Mohana Dave}, \bibinfo{person}{Chintak Mandlia}, {and}
  \bibinfo{person}{Aditya Patel}.} \bibinfo{year}{2019}\natexlab{}.
\newblock \showarticletitle{Overview of the HASOC Track at FIRE 2019: Hate
  Speech and Offensive Content Identification in Indo-European Languages}. In
  \bibinfo{booktitle}{\emph{Proceedings of the 11th Forum for Information
  Retrieval Evaluation}} (Kolkata, India) \emph{(\bibinfo{series}{FIRE '19})}.
  \bibinfo{publisher}{Association for Computing Machinery},
  \bibinfo{address}{New York, NY, USA}, \bibinfo{pages}{14–17}.
\newblock
\showISBNx{9781450377508}
\urldef\tempurl%
\url{https://doi.org/10.1145/3368567.3368584}
\showDOI{\tempurl}


\bibitem[Masud et~al\mbox{.}(2022)]%
        {sarahkdd22}
\bibfield{author}{\bibinfo{person}{Sarah Masud}, \bibinfo{person}{Manjot Bedi},
  \bibinfo{person}{Mohammad~Aflah Khan}, \bibinfo{person}{Md~Shad Akhtar},
  {and} \bibinfo{person}{Tanmoy Chakraborty}.} \bibinfo{year}{2022}\natexlab{}.
\newblock \showarticletitle{Proactively Reducing the Hate Intensity of Online
  Posts via Hate Speech Normalization}. In
  \bibinfo{booktitle}{\emph{Proceedings of the 28th ACM SIGKDD Conference on
  Knowledge Discovery and Data Mining}} (Washington DC, USA)
  \emph{(\bibinfo{series}{KDD '22})}. \bibinfo{publisher}{Association for
  Computing Machinery}, \bibinfo{address}{New York, NY, USA},
  \bibinfo{pages}{3524–3534}.
\newblock
\showISBNx{9781450393850}
\urldef\tempurl%
\url{https://doi.org/10.1145/3534678.3539161}
\showDOI{\tempurl}


\bibitem[Masud et~al\mbox{.}(2021)]%
        {masud2021hate}
\bibfield{author}{\bibinfo{person}{Sarah Masud}, \bibinfo{person}{Subhabrata
  Dutta}, \bibinfo{person}{Sakshi Makkar}, \bibinfo{person}{Chhavi Jain},
  \bibinfo{person}{Vikram Goyal}, \bibinfo{person}{Amitava Das}, {and}
  \bibinfo{person}{Tanmoy Chakraborty}.} \bibinfo{year}{2021}\natexlab{}.
\newblock \showarticletitle{Hate is the New Infodemic: A Topic-aware Modeling
  of Hate Speech Diffusion on Twitter}. In \bibinfo{booktitle}{\emph{2021 IEEE
  37th International Conference on Data Engineering (ICDE)}}.
  \bibinfo{pages}{504--515}.
\newblock
\urldef\tempurl%
\url{https://doi.org/10.1109/ICDE51399.2021.00050}
\showDOI{\tempurl}


\bibitem[Mathew et~al\mbox{.}(2019)]%
        {mathew2019spread}
\bibfield{author}{\bibinfo{person}{Binny Mathew}, \bibinfo{person}{Ritam Dutt},
  \bibinfo{person}{Pawan Goyal}, {and} \bibinfo{person}{Animesh Mukherjee}.}
  \bibinfo{year}{2019}\natexlab{}.
\newblock \showarticletitle{Spread of Hate Speech in Online Social Media}. In
  \bibinfo{booktitle}{\emph{Proceedings of the 10th ACM Conference on Web
  Science}} (Boston, Massachusetts, USA) \emph{(\bibinfo{series}{WebSci '19})}.
  \bibinfo{publisher}{Association for Computing Machinery},
  \bibinfo{address}{New York, NY, USA}, \bibinfo{pages}{173–182}.
\newblock
\showISBNx{9781450362023}
\urldef\tempurl%
\url{https://doi.org/10.1145/3292522.3326034}
\showDOI{\tempurl}


\bibitem[Mathew et~al\mbox{.}(2020)]%
        {10.1145/3415163}
\bibfield{author}{\bibinfo{person}{Binny Mathew}, \bibinfo{person}{Anurag
  Illendula}, \bibinfo{person}{Punyajoy Saha}, \bibinfo{person}{Soumya Sarkar},
  \bibinfo{person}{Pawan Goyal}, {and} \bibinfo{person}{Animesh Mukherjee}.}
  \bibinfo{year}{2020}\natexlab{}.
\newblock \showarticletitle{Hate Begets Hate: A Temporal Study of Hate Speech}.
\newblock \bibinfo{journal}{\emph{Proc. ACM Hum.-Comput. Interact.}}
  \bibinfo{volume}{4}, \bibinfo{number}{CSCW2}, Article \bibinfo{articleno}{92}
  (\bibinfo{date}{oct} \bibinfo{year}{2020}), \bibinfo{numpages}{24}~pages.
\newblock
\urldef\tempurl%
\url{https://doi.org/10.1145/3415163}
\showDOI{\tempurl}


\bibitem[Mozafari et~al\mbox{.}(2020)]%
        {10.1007/978-3-030-36687-2_77}
\bibfield{author}{\bibinfo{person}{Marzieh Mozafari}, \bibinfo{person}{Reza
  Farahbakhsh}, {and} \bibinfo{person}{No{\"e}l Crespi}.}
  \bibinfo{year}{2020}\natexlab{}.
\newblock \showarticletitle{A BERT-Based Transfer Learning Approach for Hate
  Speech Detection in Online Social Media}. In
  \bibinfo{booktitle}{\emph{Complex Networks and Their Applications VIII}},
  \bibfield{editor}{\bibinfo{person}{Hocine Cherifi}, \bibinfo{person}{Sabrina
  Gaito}, \bibinfo{person}{Jos{\'e}~Fernendo Mendes}, \bibinfo{person}{Esteban
  Moro}, {and} \bibinfo{person}{Luis~Mateus Rocha}} (Eds.).
  \bibinfo{publisher}{Springer International Publishing},
  \bibinfo{address}{Cham}, \bibinfo{pages}{928--940}.
\newblock
\showISBNx{978-3-030-36687-2}


\bibitem[Mulki et~al\mbox{.}(2019)]%
        {mulki-etal-2019-l}
\bibfield{author}{\bibinfo{person}{Hala Mulki}, \bibinfo{person}{Hatem Haddad},
  \bibinfo{person}{Chedi Bechikh~Ali}, {and} \bibinfo{person}{Halima
  Alshabani}.} \bibinfo{year}{2019}\natexlab{}.
\newblock \showarticletitle{{L}-{HSAB}: A {L}evantine {T}witter Dataset for
  Hate Speech and Abusive Language}. In \bibinfo{booktitle}{\emph{Proceedings
  of the Third Workshop on Abusive Language Online}}.
  \bibinfo{publisher}{Association for Computational Linguistics},
  \bibinfo{address}{Florence, Italy}, \bibinfo{pages}{111--118}.
\newblock
\urldef\tempurl%
\url{https://doi.org/10.18653/v1/W19-3512}
\showDOI{\tempurl}


\bibitem[Nielsen(2019)]%
        {e21050485}
\bibfield{author}{\bibinfo{person}{Frank Nielsen}.}
  \bibinfo{year}{2019}\natexlab{}.
\newblock \showarticletitle{On the Jensen–Shannon Symmetrization of Distances
  Relying on Abstract Means}.
\newblock \bibinfo{journal}{\emph{Entropy}} \bibinfo{volume}{21},
  \bibinfo{number}{5} (\bibinfo{year}{2019}).
\newblock
\showISSN{1099-4300}
\urldef\tempurl%
\url{https://doi.org/10.3390/e21050485}
\showDOI{\tempurl}


\bibitem[Pavlopoulos et~al\mbox{.}(2017)]%
        {pavlopoulos-etal-2017-deep}
\bibfield{author}{\bibinfo{person}{John Pavlopoulos},
  \bibinfo{person}{Prodromos Malakasiotis}, {and} \bibinfo{person}{Ion
  Androutsopoulos}.} \bibinfo{year}{2017}\natexlab{}.
\newblock \showarticletitle{Deep Learning for User Comment Moderation}. In
  \bibinfo{booktitle}{\emph{Proceedings of the First Workshop on Abusive
  Language Online}}. \bibinfo{publisher}{Association for Computational
  Linguistics}, \bibinfo{address}{Vancouver, BC, Canada},
  \bibinfo{pages}{25--35}.
\newblock
\urldef\tempurl%
\url{https://doi.org/10.18653/v1/W17-3004}
\showDOI{\tempurl}


\bibitem[Pedregosa et~al\mbox{.}(2011)]%
        {scikit-learn}
\bibfield{author}{\bibinfo{person}{F. Pedregosa}, \bibinfo{person}{G.
  Varoquaux}, \bibinfo{person}{A. Gramfort}, \bibinfo{person}{V. Michel},
  \bibinfo{person}{B. Thirion}, \bibinfo{person}{O. Grisel},
  \bibinfo{person}{M. Blondel}, \bibinfo{person}{P. Prettenhofer},
  \bibinfo{person}{R. Weiss}, \bibinfo{person}{V. Dubourg}, \bibinfo{person}{J.
  Vanderplas}, \bibinfo{person}{A. Passos}, \bibinfo{person}{D. Cournapeau},
  \bibinfo{person}{M. Brucher}, \bibinfo{person}{M. Perrot}, {and}
  \bibinfo{person}{E. Duchesnay}.} \bibinfo{year}{2011}\natexlab{}.
\newblock \showarticletitle{Scikit-learn: Machine Learning in {P}ython}.
\newblock \bibinfo{journal}{\emph{Journal of Machine Learning Research}}
  \bibinfo{volume}{12} (\bibinfo{year}{2011}), \bibinfo{pages}{2825--2830}.
\newblock


\bibitem[Pennington et~al\mbox{.}(2014)]%
        {pennington-etal-2014-glove}
\bibfield{author}{\bibinfo{person}{Jeffrey Pennington},
  \bibinfo{person}{Richard Socher}, {and} \bibinfo{person}{Christopher
  Manning}.} \bibinfo{year}{2014}\natexlab{}.
\newblock \showarticletitle{{G}lo{V}e: Global Vectors for Word Representation}.
  In \bibinfo{booktitle}{\emph{Proceedings of the 2014 Conference on Empirical
  Methods in Natural Language Processing ({EMNLP})}}.
  \bibinfo{publisher}{Association for Computational Linguistics},
  \bibinfo{address}{Doha, Qatar}, \bibinfo{pages}{1532--1543}.
\newblock
\urldef\tempurl%
\url{https://doi.org/10.3115/v1/D14-1162}
\showDOI{\tempurl}


\bibitem[Poletto et~al\mbox{.}(2020)]%
        {Poletto2020}
\bibfield{author}{\bibinfo{person}{Fabio Poletto}, \bibinfo{person}{Valerio
  Basile}, \bibinfo{person}{Manuela Sanguinetti}, \bibinfo{person}{Cristina
  Bosco}, {and} \bibinfo{person}{Viviana Patti}.}
  \bibinfo{year}{2020}\natexlab{}.
\newblock \showarticletitle{Resources and benchmark corpora for hate speech
  detection: a systematic review}.
\newblock \bibinfo{journal}{\emph{Language Resources and Evaluation}}
  \bibinfo{volume}{55}, \bibinfo{number}{2} (\bibinfo{date}{Sept.}
  \bibinfo{year}{2020}), \bibinfo{pages}{477--523}.
\newblock
\urldef\tempurl%
\url{https://doi.org/10.1007/s10579-020-09502-8}
\showDOI{\tempurl}


\bibitem[Qian et~al\mbox{.}(2018)]%
        {qian-etal-2018-leveraging}
\bibfield{author}{\bibinfo{person}{Jing Qian}, \bibinfo{person}{Mai ElSherief},
  \bibinfo{person}{Elizabeth Belding}, {and} \bibinfo{person}{William~Yang
  Wang}.} \bibinfo{year}{2018}\natexlab{}.
\newblock \showarticletitle{Leveraging Intra-User and Inter-User Representation
  Learning for Automated Hate Speech Detection}. In
  \bibinfo{booktitle}{\emph{Proceedings of the 2018 Conference of the North
  {A}merican Chapter of the Association for Computational Linguistics: Human
  Language Technologies, Volume 2 (Short Papers)}}.
  \bibinfo{publisher}{Association for Computational Linguistics},
  \bibinfo{address}{New Orleans, Louisiana}, \bibinfo{pages}{118--123}.
\newblock
\urldef\tempurl%
\url{https://doi.org/10.18653/v1/N18-2019}
\showDOI{\tempurl}


\bibitem[Qian et~al\mbox{.}(2021)]%
        {qian-etal-2021-lifelong}
\bibfield{author}{\bibinfo{person}{Jing Qian}, \bibinfo{person}{Hong Wang},
  \bibinfo{person}{Mai ElSherief}, {and} \bibinfo{person}{Xifeng Yan}.}
  \bibinfo{year}{2021}\natexlab{}.
\newblock \showarticletitle{Lifelong Learning of Hate Speech Classification on
  Social Media}. In \bibinfo{booktitle}{\emph{Proceedings of the 2021
  Conference of the North American Chapter of the Association for Computational
  Linguistics: Human Language Technologies}}. \bibinfo{publisher}{Association
  for Computational Linguistics}, \bibinfo{address}{Online},
  \bibinfo{pages}{2304--2314}.
\newblock
\urldef\tempurl%
\url{https://doi.org/10.18653/v1/2021.naacl-main.183}
\showDOI{\tempurl}


\bibitem[Reimers and Gurevych(2019)]%
        {reimers-gurevych-2019-sentence}
\bibfield{author}{\bibinfo{person}{Nils Reimers} {and} \bibinfo{person}{Iryna
  Gurevych}.} \bibinfo{year}{2019}\natexlab{}.
\newblock \showarticletitle{Sentence-{BERT}: Sentence Embeddings using
  {S}iamese {BERT}-Networks}. In \bibinfo{booktitle}{\emph{Proceedings of the
  2019 Conference on Empirical Methods in Natural Language Processing and the
  9th International Joint Conference on Natural Language Processing
  (EMNLP-IJCNLP)}}. \bibinfo{publisher}{Association for Computational
  Linguistics}, \bibinfo{address}{Hong Kong, China},
  \bibinfo{pages}{3982--3992}.
\newblock
\urldef\tempurl%
\url{https://doi.org/10.18653/v1/D19-1410}
\showDOI{\tempurl}


\bibitem[Ribeiro et~al\mbox{.}(2018)]%
        {ribeiro2018characterizing}
\bibfield{author}{\bibinfo{person}{Manoel Ribeiro}, \bibinfo{person}{Pedro
  Calais}, \bibinfo{person}{Yuri Santos}, \bibinfo{person}{Virgílio Almeida},
  {and} \bibinfo{person}{Wagner Meira~Jr.}} \bibinfo{year}{2018}\natexlab{}.
\newblock \showarticletitle{Characterizing and Detecting Hateful Users on
  Twitter}.
\newblock \bibinfo{journal}{\emph{Proceedings of the International AAAI
  Conference on Web and Social Media}} \bibinfo{volume}{12},
  \bibinfo{number}{1} (\bibinfo{date}{Jun.} \bibinfo{year}{2018}).
\newblock
\urldef\tempurl%
\url{https://ojs.aaai.org/index.php/ICWSM/article/view/15057}
\showURL{%
\tempurl}


\bibitem[Rizwan et~al\mbox{.}(2020)]%
        {rizwan-etal-2020-hate}
\bibfield{author}{\bibinfo{person}{Hammad Rizwan},
  \bibinfo{person}{Muhammad~Haroon Shakeel}, {and} \bibinfo{person}{Asim
  Karim}.} \bibinfo{year}{2020}\natexlab{}.
\newblock \showarticletitle{Hate-Speech and Offensive Language Detection in
  {R}oman {U}rdu}. In \bibinfo{booktitle}{\emph{Proceedings of the 2020
  Conference on Empirical Methods in Natural Language Processing (EMNLP)}}.
  \bibinfo{publisher}{Association for Computational Linguistics},
  \bibinfo{address}{Online}, \bibinfo{pages}{2512--2522}.
\newblock
\urldef\tempurl%
\url{https://doi.org/10.18653/v1/2020.emnlp-main.197}
\showDOI{\tempurl}


\bibitem[Romim et~al\mbox{.}(2021)]%
        {Romim2021}
\bibfield{author}{\bibinfo{person}{Nauros Romim}, \bibinfo{person}{Mosahed
  Ahmed}, \bibinfo{person}{Hriteshwar Talukder}, {and}
  \bibinfo{person}{Md.~Saiful Islam}.} \bibinfo{year}{2021}\natexlab{}.
\newblock \showarticletitle{Hate Speech Detection in the Bengali Language: A
  Dataset and Its Baseline Evaluation}.
\newblock In \bibinfo{booktitle}{\emph{Algorithms for Intelligent Systems}}.
  \bibinfo{publisher}{Springer Singapore}, \bibinfo{pages}{457--468}.
\newblock
\urldef\tempurl%
\url{https://doi.org/10.1007/978-981-16-0586-4_37}
\showDOI{\tempurl}


\bibitem[Rottger et~al\mbox{.}(2022)]%
        {rottger-etal-2022-two}
\bibfield{author}{\bibinfo{person}{Paul Rottger}, \bibinfo{person}{Bertie
  Vidgen}, \bibinfo{person}{Dirk Hovy}, {and} \bibinfo{person}{Janet
  Pierrehumbert}.} \bibinfo{year}{2022}\natexlab{}.
\newblock \showarticletitle{Two Contrasting Data Annotation Paradigms for
  Subjective {NLP} Tasks}. In \bibinfo{booktitle}{\emph{Proceedings of the 2022
  Conference of the North American Chapter of the Association for Computational
  Linguistics: Human Language Technologies}}. \bibinfo{publisher}{Association
  for Computational Linguistics}, \bibinfo{address}{Seattle, United States},
  \bibinfo{pages}{175--190}.
\newblock
\urldef\tempurl%
\url{https://doi.org/10.18653/v1/2022.naacl-main.13}
\showDOI{\tempurl}


\bibitem[R{\"o}ttger et~al\mbox{.}(2021)]%
        {rottger-etal-2021-hatecheck}
\bibfield{author}{\bibinfo{person}{Paul R{\"o}ttger}, \bibinfo{person}{Bertie
  Vidgen}, \bibinfo{person}{Dong Nguyen}, \bibinfo{person}{Zeerak Waseem},
  \bibinfo{person}{Helen Margetts}, {and} \bibinfo{person}{Janet
  Pierrehumbert}.} \bibinfo{year}{2021}\natexlab{}.
\newblock \showarticletitle{{H}ate{C}heck: Functional Tests for Hate Speech
  Detection Models}. \bibinfo{publisher}{Association for Computational
  Linguistics}, \bibinfo{address}{Online}, \bibinfo{pages}{41--58}.
\newblock
\urldef\tempurl%
\url{https://doi.org/10.18653/v1/2021.acl-long.4}
\showDOI{\tempurl}


\bibitem[Schmaltz(2021)]%
        {schmaltz-2021-detecting}
\bibfield{author}{\bibinfo{person}{Allen Schmaltz}.}
  \bibinfo{year}{2021}\natexlab{}.
\newblock \showarticletitle{Detecting Local Insights from Global Labels:
  Supervised and Zero-Shot Sequence Labeling via a Convolutional
  Decomposition}.
\newblock \bibinfo{journal}{\emph{Computational Linguistics}}
  \bibinfo{volume}{47}, \bibinfo{number}{4} (\bibinfo{date}{Dec.}
  \bibinfo{year}{2021}), \bibinfo{pages}{729--773}.
\newblock
\urldef\tempurl%
\url{https://doi.org/10.1162/coli_a_00416}
\showDOI{\tempurl}


\bibitem[Schmidhuber(1997)]%
        {schmidhuber1997long}
\bibfield{author}{\bibinfo{person}{J{\"u}rgen Schmidhuber}.}
  \bibinfo{year}{1997}\natexlab{}.
\newblock \showarticletitle{Long Short-Term Memory}.
\newblock \bibinfo{journal}{\emph{Neural Computation}} \bibinfo{volume}{9},
  \bibinfo{number}{8} (\bibinfo{year}{1997}), \bibinfo{pages}{1735--1780}.
\newblock


\bibitem[Schmidt and Wiegand(2017)]%
        {schmidt-wiegand-2017-survey}
\bibfield{author}{\bibinfo{person}{Anna Schmidt} {and} \bibinfo{person}{Michael
  Wiegand}.} \bibinfo{year}{2017}\natexlab{}.
\newblock \showarticletitle{A Survey on Hate Speech Detection using Natural
  Language Processing}. In \bibinfo{booktitle}{\emph{Proceedings of the Fifth
  International Workshop on Natural Language Processing for Social Media}}.
  \bibinfo{publisher}{Association for Computational Linguistics},
  \bibinfo{address}{Valencia, Spain}, \bibinfo{pages}{1--10}.
\newblock
\urldef\tempurl%
\url{https://doi.org/10.18653/v1/W17-1101}
\showDOI{\tempurl}


\bibitem[Silva et~al\mbox{.}(2021)]%
        {Silva_Mondal_Correa_Benevenuto_Weber_2021}
\bibfield{author}{\bibinfo{person}{Leandro Silva}, \bibinfo{person}{Mainack
  Mondal}, \bibinfo{person}{Denzil Correa}, \bibinfo{person}{Fabrício
  Benevenuto}, {and} \bibinfo{person}{Ingmar Weber}.}
  \bibinfo{year}{2021}\natexlab{}.
\newblock \showarticletitle{Analyzing the Targets of Hate in Online Social
  Media}.
\newblock \bibinfo{journal}{\emph{Proceedings of the International AAAI
  Conference on Web and Social Media}} \bibinfo{volume}{10},
  \bibinfo{number}{1} (\bibinfo{date}{Aug.} \bibinfo{year}{2021}),
  \bibinfo{pages}{687--690}.
\newblock
\urldef\tempurl%
\url{https://ojs.aaai.org/index.php/ICWSM/article/view/14811}
\showURL{%
\tempurl}


\bibitem[Sue(2010)]%
        {sue2010microaggressions}
\bibfield{author}{\bibinfo{person}{Derald~Wing Sue}.}
  \bibinfo{year}{2010}\natexlab{}.
\newblock \bibinfo{booktitle}{\emph{Microaggressions in everyday life: Race,
  gender, and sexual orientation}}.
\newblock \bibinfo{publisher}{John Wiley \& Sons}.
\newblock


\bibitem[Toraman et~al\mbox{.}(2022)]%
        {toraman2022largescale}
\bibfield{author}{\bibinfo{person}{Cagri Toraman}, \bibinfo{person}{Furkan
  Şahinuç}, {and} \bibinfo{person}{Eyup~Halit Yılmaz}.}
  \bibinfo{year}{2022}\natexlab{}.
\newblock \bibinfo{title}{Large-Scale Hate Speech Detection with Cross-Domain
  Transfer}.
\newblock
\newblock
\showeprint[arxiv]{2203.01111}~[cs.CL]


\bibitem[Vaswani et~al\mbox{.}(2017)]%
        {NIPS2017_3f5ee243}
\bibfield{author}{\bibinfo{person}{Ashish Vaswani}, \bibinfo{person}{Noam
  Shazeer}, \bibinfo{person}{Niki Parmar}, \bibinfo{person}{Jakob Uszkoreit},
  \bibinfo{person}{Llion Jones}, \bibinfo{person}{Aidan~N Gomez},
  \bibinfo{person}{\L~ukasz Kaiser}, {and} \bibinfo{person}{Illia Polosukhin}.}
  \bibinfo{year}{2017}\natexlab{}.
\newblock \showarticletitle{Attention is All you Need}. In
  \bibinfo{booktitle}{\emph{Advances in Neural Information Processing
  Systems}}, \bibfield{editor}{\bibinfo{person}{I.~Guyon},
  \bibinfo{person}{U.~Von Luxburg}, \bibinfo{person}{S.~Bengio},
  \bibinfo{person}{H.~Wallach}, \bibinfo{person}{R.~Fergus},
  \bibinfo{person}{S.~Vishwanathan}, {and} \bibinfo{person}{R.~Garnett}}
  (Eds.), Vol.~\bibinfo{volume}{30}. \bibinfo{publisher}{Curran Associates,
  Inc.}
\newblock
\urldef\tempurl%
\url{https://proceedings.neurips.cc/paper/2017/file/3f5ee243547dee91fbd053c1c4a845aa-Paper.pdf}
\showURL{%
\tempurl}


\bibitem[Warner and Hirschberg(2012)]%
        {warner-hirschberg-2012-detecting}
\bibfield{author}{\bibinfo{person}{William Warner} {and} \bibinfo{person}{Julia
  Hirschberg}.} \bibinfo{year}{2012}\natexlab{}.
\newblock \showarticletitle{Detecting Hate Speech on the World Wide Web}. In
  \bibinfo{booktitle}{\emph{Proceedings of the Second Workshop on Language in
  Social Media}}. \bibinfo{publisher}{Association for Computational
  Linguistics}, \bibinfo{address}{Montr{\'e}al, Canada},
  \bibinfo{pages}{19--26}.
\newblock
\urldef\tempurl%
\url{https://aclanthology.org/W12-2103}
\showURL{%
\tempurl}


\bibitem[Waseem and Hovy(2016)]%
        {waseem-hovy-2016-hateful}
\bibfield{author}{\bibinfo{person}{Zeerak Waseem} {and} \bibinfo{person}{Dirk
  Hovy}.} \bibinfo{year}{2016}\natexlab{}.
\newblock \showarticletitle{Hateful Symbols or Hateful People? Predictive
  Features for Hate Speech Detection on {T}witter}. In
  \bibinfo{booktitle}{\emph{Proceedings of the {NAACL} Student Research
  Workshop}}. \bibinfo{publisher}{Association for Computational Linguistics},
  \bibinfo{address}{San Diego, California}, \bibinfo{pages}{88--93}.
\newblock
\urldef\tempurl%
\url{https://doi.org/10.18653/v1/N16-2013}
\showDOI{\tempurl}


\bibitem[Wolf et~al\mbox{.}(2020)]%
        {wolf-etal-2020-transformers}
\bibfield{author}{\bibinfo{person}{Thomas Wolf}, \bibinfo{person}{Lysandre
  Debut}, \bibinfo{person}{Victor Sanh}, \bibinfo{person}{Julien Chaumond},
  \bibinfo{person}{Clement Delangue}, \bibinfo{person}{Anthony Moi},
  \bibinfo{person}{Pierric Cistac}, \bibinfo{person}{Tim Rault},
  \bibinfo{person}{Remi Louf}, \bibinfo{person}{Morgan Funtowicz},
  \bibinfo{person}{Joe Davison}, \bibinfo{person}{Sam Shleifer},
  \bibinfo{person}{Patrick von Platen}, \bibinfo{person}{Clara Ma},
  \bibinfo{person}{Yacine Jernite}, \bibinfo{person}{Julien Plu},
  \bibinfo{person}{Canwen Xu}, \bibinfo{person}{Teven Le~Scao},
  \bibinfo{person}{Sylvain Gugger}, \bibinfo{person}{Mariama Drame},
  \bibinfo{person}{Quentin Lhoest}, {and} \bibinfo{person}{Alexander Rush}.}
  \bibinfo{year}{2020}\natexlab{}.
\newblock \showarticletitle{Transformers: State-of-the-Art Natural Language
  Processing}. In \bibinfo{booktitle}{\emph{Proceedings of the 2020 Conference
  on Empirical Methods in Natural Language Processing: System Demonstrations}}.
  \bibinfo{publisher}{Association for Computational Linguistics},
  \bibinfo{address}{Online}, \bibinfo{pages}{38--45}.
\newblock
\urldef\tempurl%
\url{https://doi.org/10.18653/v1/2020.emnlp-demos.6}
\showDOI{\tempurl}


\bibitem[Zampieri et~al\mbox{.}(2019a)]%
        {zampierietal2019}
\bibfield{author}{\bibinfo{person}{Marcos Zampieri}, \bibinfo{person}{Shervin
  Malmasi}, \bibinfo{person}{Preslav Nakov}, \bibinfo{person}{Sara Rosenthal},
  \bibinfo{person}{Noura Farra}, {and} \bibinfo{person}{Ritesh Kumar}.}
  \bibinfo{year}{2019}\natexlab{a}.
\newblock \showarticletitle{{Predicting the Type and Target of Offensive Posts
  in Social Media}}. In \bibinfo{booktitle}{\emph{Proceedings of NAACL}}.
\newblock


\bibitem[Zampieri et~al\mbox{.}(2019b)]%
        {zampieri-etal-2019-semeval}
\bibfield{author}{\bibinfo{person}{Marcos Zampieri}, \bibinfo{person}{Shervin
  Malmasi}, \bibinfo{person}{Preslav Nakov}, \bibinfo{person}{Sara Rosenthal},
  \bibinfo{person}{Noura Farra}, {and} \bibinfo{person}{Ritesh Kumar}.}
  \bibinfo{year}{2019}\natexlab{b}.
\newblock \showarticletitle{{S}em{E}val-2019 Task 6: Identifying and
  Categorizing Offensive Language in Social Media ({O}ffens{E}val)}. In
  \bibinfo{booktitle}{\emph{Proceedings of the 13th International Workshop on
  Semantic Evaluation}}. \bibinfo{publisher}{Association for Computational
  Linguistics}, \bibinfo{address}{Minneapolis, Minnesota, USA},
  \bibinfo{pages}{75--86}.
\newblock
\urldef\tempurl%
\url{https://doi.org/10.18653/v1/S19-2010}
\showDOI{\tempurl}


\bibitem[Zhou et~al\mbox{.}(2021)]%
        {zhou-etal-2021-hate}
\bibfield{author}{\bibinfo{person}{Xianbing Zhou}, \bibinfo{person}{Yang Yong},
  \bibinfo{person}{Xiaochao Fan}, \bibinfo{person}{Ge Ren},
  \bibinfo{person}{Yunfeng Song}, \bibinfo{person}{Yufeng Diao},
  \bibinfo{person}{Liang Yang}, {and} \bibinfo{person}{Hongfei Lin}.}
  \bibinfo{year}{2021}\natexlab{}.
\newblock \showarticletitle{Hate Speech Detection Based on Sentiment Knowledge
  Sharing}. In \bibinfo{booktitle}{\emph{Proceedings of the 59th Annual Meeting
  of the Association for Computational Linguistics and the 11th International
  Joint Conference on Natural Language Processing (Volume 1: Long Papers)}}.
  \bibinfo{publisher}{Association for Computational Linguistics},
  \bibinfo{address}{Online}, \bibinfo{pages}{7158--7166}.
\newblock
\urldef\tempurl%
\url{https://doi.org/10.18653/v1/2021.acl-long.556}
\showDOI{\tempurl}


\end{thebibliography}

\newpage
\appendix
\section*{Appendix}

\section{Limitations}
\label{app:lim}
Given the subjective nature and lack of standard definitions for hate, collecting and tagging large-scale datasets for hate speech classification is difficult. We observed the same while compiling \dataset. At the same time, due to neutral seeding and lack of pervasive lexicons, the hateful samples obtained in \dataset\ are similar in context and syntax to the non-hateful ones. It, in turn, leads to higher annotation disagreements and a higher misclassification rate when modeling. Further, with scale, one has to move away from an expert annotation setup to a crowdsourced design, leading to a loss of expertise in place of the wisdom of the masses. \citet{rottger-etal-2022-two} presents a spectrum of data annotation paradigms for NLP that can help assuage some of these concerns in the future. At the same time, one must acknowledge the annotator's bias (irrespective of the annotation setup). Our dataset, too, suffers from mislabelling arising from annotator bias and disagreements (we highlight one such example during error analysis in Section \ref{sec:error_analysis}). In addition to the limitations arising from the dataset, we also have limits at the modeling level. To effectively utilize our method, it is required to have access to historical and topological information.

For future work, we press for a more neuro-symbolic approach that facilitates the integration of commonsense knowledge, factual analysis, and stereotyping motives for better hate detection. Using prompts as a source of explanation-driven classification can be another future direction.

\section{Ethical Considerations}
\label{app:ethical}
Datasets and pertained language models for hate speech can be used as a force for both good and bad. While on the one hand, they help characterize what forms of hate spread on a platform. On the other hand, nefarious fractions can also introduce harmful content on the internet. The persistence of hate speech on platforms can impact the ability of vulnerable and marginalized groups to effectively and freely communicate on the web. Thereby early flagging and moderation should be the aim. We reiterate here that we aim not to aid in spreading hate, nor do we wish any vulnerable group to be harmed or negatively affected due to this research. Instead, by providing a more nuanced real-world view of hateful discourse, we hope to shift the focus on better understanding topical and linguistic signals to provide a holistic view of hate speech. 

During the data collection and annotation phases, we continuously monitored the annotation quality and frequently incorporated feedback from the annotator to improve the guidelines. We tried to be cognizant of annotation disagreements and tried to bring everyone on the same page during the testing phase to avoid invalidating a renewed point of view. We also ensured that quality time was provided for the labeling and that they did not feel fatigued.

\section{Benchhmark Hate Speech Datasets}
\label{app:benchmark_datasets}
An outline of the various benchmark datasets under consideration is provided in Table \ref{tab:comp_data}. For our current analysis, we omitted the neutrally seeded dataset from Stormfront \cite{de-gibert-etal-2018-hate} as it is an unmoderated platform. Additionally, it can be observed that shared task-based datasets (curated and annotated in a controlled setting) have a higher and more equitable distribution of hate labels compared to their large-scale counterparts curated in the wild. All these datasets have different sets of labels and cover varying aspects of hatefulness and offense. For the ImpHate dataset, we combined the classes of implicit and explicit hate into a single hate category. For the rest of the datasets, the label distribution stays intact. 
\begin{table}[!t]
\centering
\caption{The total number of samples, unique labels, and the percentage of hateful samples in hate datasets. The hate percentage is simply calculated as the ratio of samples marked as hate vs total samples in the dataset.}
\label{tab:comp_data}
\resizebox{\columnwidth}{!}{
\begin{tabular}{|c|c|c|c|c|c|}
\hline
\textbf{Dataset} & \textbf{\# Samples} & \textbf{\# Labels} & \textbf{Hate \%}\\ \hline
\dataset & $51367$ & $4$ (H,O,P,N) & $0.0723$\\\hdashline
Founta \cite{founta2018large} & $59189$ & $4$ (H,A,S,N) & $0.0405$\\\hdashline
Davidson \cite{founta2018large} & $24783$ & $3$ (H,O,N) & $0.0577$\\\hdashline
HASOC19 \cite{HASCO19} & $17657$ &$2$ (H,N) & $0.3492$ \\ \hdashline
OLID \cite{zampierietal2019} & $10592$ & $2$ (O,N)& $0.3291$ \\ \hline
HatEval \cite{basile-etal-2019-semeval} & $19600$ & $2$ (H,N) & $0.4188$ \\ \hdashline
ImpHate \cite{elsherief-etal-2021-latent} & $21480$ & $2$ (H,N) & $0.3812$ \\ \hdashline
RETINA \cite{masud2021hate} & $23748$ & $2$ (H,N) & $0.0468$ \\ \hline
\end{tabular}
}
\vspace{-3mm}
\end{table}

\section{Extended Experiments}
\label{app:yet_mbert}
We also extend the experiments in Section \ref{sec:yet_another} to employ mBERT. The transformer model was frozen and a classification head with the required number of class labels was employed for the respective experiment. Table \ref{tab:mdd_all_datasets} highlights capturing the inter-class similarity via maximum mean discrepancy (MMD) \cite{mmd}. Meanwhile Table \ref{tab:hate_perf_mbert} employs mBERT to perform within dataset as well as data validation (\dataset with others) classification. Lastly, Table \ref{tab:cross_perf_mbert} records the cross-dataset generalizability among the $3$ Hindi/Hindi-codemixed datasets. We observe similar patterns across the $4$ experiments as detailed in Section \ref{sec:yet_another}.  
\begin{table}[!h]
% \vspace{-3mm}
\centering
\caption{Inter-class similarity within a hate dataset measured via maximum mean discrepancy (MMD) obtained via mBERT embeddings. The lower the MMD, the harder it will be to separate the classes. For each dataset, we report the median (mean) MMD over 100 runs with 30 random pair (label-wise) of samples (without replacement).}
\label{tab:mdd_all_datasets}
\resizebox{\columnwidth}{!}{
\begin{tabular}{|c|c|c|c||c|c|c|c|c|}
\hline
\textbf{Dataset} & \multicolumn{2}{c|}{\textbf{Label}} & \textbf{JS} & \textbf{Dataset} & \multicolumn{2}{c|}{\textbf{Label}} & \textbf{JS}  \\\hline
\multirow{6}{*}{\dataset}   & H & O & 0.037 (0.024) &     \multirow{6}{*}{Founta \cite{founta2018large}} & H & A & 0.023 (0.018) \\
                            & H & P & 0.028 (0.018) &                             & H & S & 0.061 (0.051) \\
                            & H & N & 0.030 (0.022) &                             & H & N & 0.044 (0.031) \\
                            & O & P & 0.032 (0.022) &                             & A & S & 0.072 (0.064) \\
                            & O & N & 0.027 (0.020) &                             & A & N & 0.060 (0.050)\\
                            & P & N & 0.023 (0.016) &                             & S & N & 0.035 (0.031)\\ \hdashline
\multirow{3}{*}{Davidson \cite{davidson2017automated}}   & H & O & 0.042 (0.029) & HASOC19 \cite{HASCO19}                     & H & N & 0.108 (0.096)\\  \cdashline{5-8}
                            & H & N & 0.039 (0.025) & OLID  \cite{zampierietal2019}                        & O & N & 0.036 (0.021) \\ \cdashline{5-8}
                            & O & N & 0.069 (0.051) & HatEval \cite{basile-etal-2019-semeval}                    & H & N & 0.039 (0.039) \\ \hdashline
ImpHate \cite{elsherief-etal-2021-latent}                     & H & N & 0.044 (0.034) & RETINA \cite{masud2021hate}                       & H & N  & 0.043 (0.031) \\ \hline
\end{tabular}
}
\vspace{-3mm}
\end{table}

\begin{table} [!h]
\caption{Cross-dataset performance comparison among \dataset, HASOC19 and RETINA with mBERT as base model. We report ROC-AUC (Matthews correlation coefficient) on binarised labels.}
\label{tab:cross_perf_mbert}
\resizebox{\columnwidth}{!}{
\begin{tabular}{|c|*{3}{c|}}
\hline
\textbf{Train$\downarrow$Test$\rightarrow$} & \dataset & RETINA \cite{masud2021hate} & HASOC19 \cite{HASCO19} \\\hline
\dataset & - & $0.609$ ($0.124$) & $0.701$ ($0.371$) \\ \hdashline
RETINA \cite{masud2021hate} & $0.503$ ($0.059$)  & - & $0.502$ ($0.040$)  \\ \hdashline
HASOC19 \cite{HASCO19} & $0.516$ ($0.058$)  & $0.525$ ($0.058$) & - \\\hline 
\end{tabular}
}
\vspace{-3mm}
\end{table}

\begin{table}[]
\caption{Performance comparison of hate speech datasets. We report the overall macro-F1 and Matthews correlation coefficient (MCC) for: a) Within dataset training and testing to predict one of the class labels of the dataset. b) Cross-dataset training and testing, when the dataset is mixed with \dataset for monitoring and predicting data drift. Here C\textsubscript{(1,2)} represent the combined macro-F1 for cases (a) and (b), respectively.}
\label{tab:hate_perf_mbert}
\resizebox{\columnwidth}{!}{
\begin{tabular}{|c|c|c|}
\hline
\textbf{Dataset} & \textbf{C\textsubscript{1} (MCC)} & \textbf{C\textsubscript{2} (MCC)}\\\hdashline
\dataset & $0.28$ ($0.139$) & - \\ \hdashline
Founta \cite{founta2018large} & $0.51$ ($0.546$) & $0.94$ ($0.855$)  \\ \hdashline
Davidson \cite{davidson2017automated} & $0.51$ ($0.493$)& $0.98$ ($0.961$)  \\ \hdashline
HASOC19 \cite{HASCO19} & $0.71$ ($0.427$) & $0.81$ ($0.956$)  \\ \hdashline
OLID \cite{zampierietal2019} & $0.62$ ($0.286$) & $0.91$ ($0.888$)  \\ \hdashline
HatEval \cite{basile-etal-2019-semeval} & $0.58$ ($0.180$) & $0.894$ ($0.811$) \\ \hdashline
ImpHate \cite{elsherief-etal-2021-latent} & $0.66$ ($0.330$) & $0.98$ ($0.331$)  \\ \hdashline
RETINA \cite{masud2021hate} & $0.49$ & $0.99$ ($0.99$) \\ \hline
\end{tabular}
}
\vspace{-4mm}
\end{table}

\section{Annotator Details}
\label{app:anno_pipe}
While their political views vary, none of the annotators in any group are associated with any political organization. \textit{Group A} consists of three female annotators aged between $24-34$. \textit{Group B} annotators were of Indian origin within the age range $20$-$64$ years. They all knew the socio-political discourse in USA, UK, and India and were fluent in code-mixed Hindi. All had prior experience using social media and annotating textual datasets. $6$ annotators identified as females and $4$ identified as males. $6$ annotators worked on English and Hinglish data, $3$ worked on Hindi (Devanagiri), and $1$ worked on both. The crowdsourced annotators were compensated \$$1$  per tweet and were required to annotate $500$ tweets per week.  

\begin{table*}[]
\caption{Sample tweets (verbatim) with their annotated labels and reasoning for annotation.}
\label{tab:anno_reason}
\resizebox{\textwidth}{!}{
\begin{tabular}{|p{30em}|c|p{30em}|}
\hline
\textbf{Sample} &
  \textbf{Label} &
  \textbf{Rationale} \\\hline
'Unko aukaat dikhaenge' (\small Will show them their place), had said a young man with face covered amidst chanting Jai Shri Ram. Northeast \#Delhi saw nothing but communal riot. People killed, shops and vehicles burnt. \#DelhiRiots &
  H &
  Refer Hate point 1 (Distorting views of Hindu community) \\\hline
Chincommie virus killing off Chinese. @china leaders killing off Wuhan politicos. Build that wall faster @realDonaldTrump @GOP @CDCgov &
  H &
  Refer Hate point 1 (demeaning a race. While Chinese population is not a minority but w.r.t the topic of this post they are a vulnerable group. Also note that while Communist China is not hurtful , the term commie like nigger is a derogatory remarks) \\\hline
\#hindulivesmatter \#Hinduphobia Sometimes you just have to jump into activist mode esp. when \#India \& \#Hinduism are denigrated. An offensive T shirt has been removed after complaints to the T shirt company. When the next time such things happens, WILL YOU ACT ????? &
  P &
  Refer to Provocative Point 2 (An indirect call to action) \\\hline
\#unsungHindus \#Hinduunity \#hindulivesmatter Sometimes you just have to jump into activist mode especially when India and Hinduism are denigrated. Thank you Mahalakshmi Ganapathy Vijay Kumar Shourie Bannai. &
  N &
  While this post is similar to the above one, and is w.r.t to the same event. This post is not  provocative because instead of trying to stir an action it is trying to put this in a positive light (based on the available content)\\\hline
\end{tabular}
}
\end{table*}

\section{Sample Annotations}
\label{app:anno_def}
Table \ref{tab:anno_reason} lists some examples and the corresponding reasoning for the assigned label. Note that while \dataset\ is curated based on socio-political topics, the tweets obtained for a topic are not geo-tagged.

\section{Experimental Setup}
\label{app:exp_setup}
For the statistical models of Multinomial Naive Bayes, Logistic Regression, and LinearSVC, we used word TF-IDF features using the Scikit-learn \cite{scikit-learn}. Meanwhile, CNN, LSTM, and Founta baselines incorporate Glove EN+HI embeddings \cite{pennington-etal-2014-glove}. The transformer-based models of mBERT, ARHNet, HurtBERT, and \model\ were trained using the Huggingface \cite{wolf-etal-2020-transformers} in PyTorch. After manual hyperparameter tuning, we select the following values: learning rate: $2e-5$, weight decay: $1e-4$, batch size: $32$, max epochs: $20$, optimizer: Adam \cite{kingma2014adam}. We select the best validation model after the training macro-F1 has crossed $70\%$. All the codes were seeded to a value of $42$. NVIDIA RTX A6000 of $48$GB was used for all the experiments. All the transformer-based models occupied around $15-19$GB of GPU and required approximately $6-8$ minutes per epoch.

\section{Sample Exemplars}
\label{app:exemplar}
Some sample exemplars from the training and validation set are enlisted (verbatim) below:
\begin{enumerate}[noitemsep, nolistsep, leftmargin=1em]
    \item \textbf{Offensive}: "\textit{\$MENTION\$ DERANGED DELUSIONAL DUMB DICTATOR DONALD IS MENTALLY UNSTABLE! I WILL NEVER VOTE REPUBLICAN AGAIN IF THEY DON'T STAND UP TO THIS TYRENT LIVING IN THE WHITE HOUSE! fk republicans worst dictator ever unstable dictator \$URL\$}"
    \begin{itemize}[noitemsep, nolistsep, leftmargin=1em]
        \item \textbf{Offensive}: \textit{"\$MENTION\$ COULD WALK ON WATER AND THE never trump WILL CRAP ON EVERYTHING HE DOES. SHAME IN THEM. UNFOLLOW ALL OF THEM PLEASE!"} 
        \item \textbf{Offensive}: \textit{"\$MENTION\$ \$MENTION\$ \$MENTION\$ AND Remember president loco SAID MEXICO WILL PAY FUC**kfu ck trump f*** gop f*** republicans Make go fund me FOR HEALTH CARE, COLLEGE EDUCATION , CLIMATE CHANGE, SOMETHING GOOD AND POSITIVE !! Not for a f**ing wall go fund the wall the resistance resist \$URL\$"}
    \end{itemize}
    \item \textbf{Provocative}: \textit{"Where is Award Wapsi (\small return) gang? Are they hiding behind $MENTION$ Lungi (\small dress)? hindu lives matter"}    
    \begin{itemize}[noitemsep, nolistsep, leftmargin=1em]
        \item \textbf{Provocative}: \textit{"Where's the outrage? Where's the Musalman-Khatre-Mein-hai (\small Muslims-are-in-danger) Gang? Where's the Bhim army now? Pathetic cronies. hindu lives matter \$URL\$"}
        \item \textbf{Provocative}: \textit{"\$MENTION\$ Where are the Bollywood liberals ? Where is their condemnation and outrage ? hindu lives matter"}
   \end{itemize}
    \item \textbf{Hate}: \textit{"Ankit Sharma. Gathered intelligence to protect this nation. In this nation, he is dragged by 40-50 people, undressed, tortured, face and upper body smashed with heavy objects till his last breath. This is my nation. delhi riots2020 arrest tahir hussain $URL$"}
    \begin{itemize}[noitemsep, nolistsep, leftmargin=1em]
        \item \textbf{Hate}: \textit{"Appalled by the targeted killing and violence against Muslims in delhi. With so many lives lost, houses burned, and places of worship torched; we request Government of India to restore normalcy ASAP. delhi riots delhi riots2020 \$URL\$"}
        \item \textbf{Provocative}: \textit{"The son of Muddasir Khan, one of the martyrs of the anti Muslim Delhi riots, weeps over his innocent father's body. We are keeping an account. We won't forget anything. ANYTHING AT ALL! delhi riots facism indian muslims in danger pic.\$URL\$"}
    \end{itemize}
\end{enumerate}
The example $1$ from the above list is from the training data. We observe that the retrieved exemplars are high quality as we adopt a label-based exemplar search for the training samples. We query exemplars from the training dataset without knowing the label for the validation and test sets, as they are supposed to be unseen examples. Thus, this might lead to dubious retrieval results. In example $2$, we see that the model retrieves apt exemplars sharing the same theme, sentiment, and label with the root tweet. Example $3$, on the other hand, retrieves exemplars that share topical similarity with the root tweet but miss the actual context and the label.

\begin{table*}[!t]
\caption{Performance analysis of competing models. We report  class-wise and overall precision (P), recall (R) and macro-F1 (F1), and combined accuracy (Acc). }
\label{tab:perffull}
\resizebox{\textwidth}{!}{
\begin{tabular}{|c|*{16}{c|}}
\hline
\multirow{2}{*}{\textbf{Model}} & \multicolumn{3}{|c|}{\textbf{Hate}} & \multicolumn{3}{|c|}{\textbf{Offensive}} & \multicolumn{3}{|c|}{\textbf{Provocative}} & \multicolumn{3}{|c|}{\textbf{Neutral}} & \multicolumn{4}{|c|}{\textbf{Combined}}\\\cdashline{2-17}
& P & R & F1 & P & R & F1 & P & R & F1 & P & R & F1 & P & R & F1 & Acc\\\hline
NB & $0.4468$ & $0.1129$ & $0.1802$ & $0.4027$ & $0.4287$ & $0.4153$ & $0.4253$ & $0.2322$ & $0.3004$ & $0.6557$ & $0.8150$ & $0.7267$ & $0.4826$ & $0.3972$ &$0.4057$ & $0.5838$\\\hdashline

LR &  $\mathbf{0.4646}$ & $0.1239$ & $0.1957$ & $0.5714$ & $0.3618$ & $0.4430$ & $0.4764$ & $0.2490$ & $0.3271$ & $0.6576$ & $0.9004$ & $\mathbf{0.7601}$ & $\mathbf{0.5425}$ & $0.4088$ &$0.4315$ & $\mathbf{0.6258}$\\\hdashline

SVC & $0.4357$ & $0.1639$ & $0.2382$ & $0.5192$ & $0.4002$ & $0.4520$ & $0.4072$ & $0.3099$ & $0.3519$ & $0.6690$ & $0.8232$ & $0.7382$ & $0.5078$ & $0.4243$ & $0.4451$ & $0.6036$ \\\hdashline

Davidson & $0.1732$ & $0.4516$ & $0.2504$ & $0.3814$ & $0.4362$ & $0.4070$ & $0.3472$ & $0.4038$ & $0.3734$ & $0.7402$ & $0.5143$ & $0.6069$ & $0.4105$ & $0.4515$ & $0.4094$ & $0.4748$\\\hline

CNN & $0.1824$ & $0.3844$ & $0.2474$ & $0.3307$ & $\mathbf{0.5569}$ & $0.4148$ & $0.3593$ & $0.3848$ & $0.3716$ & $0.7683$ & $0.4936$ & $0.6019$ & $0.4101$ & $0.4548$ & $0.4087$ & $0.4732$\\\hdashline

LSTM & $0.2083$ & $0.3387$ & $0.2579$ & $0.4168$ & $0.4907$ & $0.4508$ & $0.3404$ & $0.5005$ & $0.4052$ & $\mathbf{0.7715}$ & $0.5398$ & $0.6352$ & $0.4347$ & $0.4674$ & $0.4373$ & $0.5094$\\\hdashline

Founta & $0.1345$ & $\mathbf{0.4543}$ & $0.2075$ & $0.3392$ & $0.3841$ & $0.3603$ & $0.2925$ & $0.2806$ & $0.2864$ & $0.7022$ & $0.4726$ & $0.5650$ & $0.3671$ & $0.3979$ & $0.3548$ & $0.4180$\\\hline

mBERT & $0.3309$ & $0.2419$ & $0.2795$ & $0.4884$ & $0.4994$ & $0.4939$ & $\mathbf{0.3964}$ & $0.3754$ & $0.3856$ & $0.7119$ & $\mathbf{0.7458}$ & $0.7285$ & $0.4819$ & $0.4656$ & $0.4719$ & $0.5945$ \\\hdashline

ARHNet & $0.2770$ & $0.2956$ & $0.2860$ 
& $0.5056$ & $0.4423$ & $0.4719$ 
& $0.3729$ & $\mathbf{0.4492}$ & $\mathbf{0.4075}$ 
& $0.7321$ & $0.6968$ & $0.7140$ 
& $0.4719$ & $0.4710$ & $0.4699$ & $0.5769$ \\\hdashline

HurtBERT & $0.2836$ & $0.2607$ & $0.2717$ 
& $\mathbf{0.5266}$ & $0.4646$ & $0.4937$ 
& $0.3751$ & $0.4341$ & $0.4024$ 
& $0.7676$ & $0.7092$ & $0.7143$ 
& $0.7194$ & $0.4672$ & $0.4705$ & $0.5818$ \\\hline

\model & $0.3698$ & $0.3091$ & $\mathbf{0.3367}$ 
& $0.5103$ & $0.4895$ & $\mathbf{0.4997}$ 
& $0.3852$ & $0.4294$ & $0.4061$ 
& $0.7392$ & $0.7323$ & $0.7358$ 
& $0.5011$ & $\mathbf{0.4901}$ & $\mathbf{0.4946}$ & $0.6013$\\\hline  %mcc 0.3450 and roc_auc 0.6649
\end{tabular}
}
% \vspace{-2.5mm}
\end{table*}

\section{Performance Comparision}
\label{app:perf}
\textbf{Incorporating Past Signals.} When reporting a tweet, Twitter asks for other similar tweets on the user's timeline that appear offensive to the reporter. Similarly, when reporting on WhatsApp, it captures the last 5 conversations between a reporter and the person being reported. Thus, we hypothesize and observe (Example 2 in Table 7) that contextual information is essential for adequately classifying hateful contexts. In this regard, a user's timeline or most recent interactions on the platform are quick to obtain. Meanwhile, network-level signals are tricky in real time and, for research purposes, are primarily stored in terms of temporal snapshots (like in our case). Interestingly we observe that the most recent timelines as small as +/-5 combined with a slightly stale but weighted network of most recent interactions, as infused in HEN-mBERT, lead to a 6-point increase in macro-F1 of hate class (compared to mBERT). In agreement with the reviewer, we hypothesize obtaining timeline interactions for real-world deployment is more plausible. One must also account for the network effect as the information a user eventually interacts with is mainly obtained from their friend circle on the platform \cite{10.1145/3415163}.

\textbf{Evaluation.} This section provides the breakdown of the fine-grained performance of our baseline models and our mixture-of-experts \model. Table \ref{tab:perffull}
further highlights the imbalance in the performance of the baseline methods, where they optimize for either precision or recall, thereby hampering the overall F1. Meanwhile, our proposed models are comparatively balanced, improving overall model performance. While our dataset is not binary, thinking in terms of hateful vs. non-hateful, from the point of view of a targetted community, it is not desirable when hate does not get flagged as it allows the hateful content to persist and spread on the system. Thus, it would be crucial to measure hate speech detection by improving contextual detection of hateful classes (H, O, P) over neutral (N). It is essential to point out that the accuracy (Acc) is not a viable metric given the high skew of labels in the \dataset\, which is also representative of the skew in other hate speech datasets. Hence in our analysis, we focus on class-wise and overall F1. As we observe in Table \ref{tab:perffull}, non-contextual embedding-based models either have higher precision or recall leading to extremely low F1 for the hate class. Meanwhile, contextual embedding-based systems, including ours, provide a more balanced precision and recall leading to higher F1. For example, we observe that the LR model (M2) has a high recall for the neutral class ($0.9$), thus, indicating that it is aggressive in classifying every tweet as non-hate (the majority class). Subsequently, it has a low recall for hate, $0.13$. Meanwhile, for HEN-mBERT, this difference in recall between H-N is comparatively lower. 

We also look into the overlap in terms of users covered in the respective classes in the annotated and predicted test samples. 
$354$ unique user ids are obtained in the hateful samples of the test set via the ground-truth annotations. HEN-mBERT in the hate class has $354$,$293$, and $158$ users in annotated, predicted, and overlapping sets. Meanwhile, the numbers for LR (M2) stand at $354$,$86$, and $48$. To simplify, the annotated ids are from the crowdsourced hateful samples (354 in this case), the predicted ones are model specific, and the overlap is the users in the predicted set which are actually present in the annotated/ground-truth (true positives). LR possesses comparatively lower user overlap due to its tendency to consider everything neutral. Meanwhile, by carrying a more balanced P-R per class, our model ensures that no group/class is penalized more than others. While our model gives more false positives for hate (174) than the LR model (38), this does not necessarily imply that our model is overfitting. The proposed model is relatively more aggressive in predicting hate than the LR model. The performance is still low for hate classification ($44.6$\% for our model and $13.5$\% for LR). In a real-world scenario, it will be better to protect the targetted groups if models are more aggressive in predicting hatefulness, as the moderators could flag the false positives in the final reporting. If the model is aggressively predicting non-hate, it defeats the very purpose of hate speech detection by exposing at-risk groups to a higher volume of hateful content.
\end{document}